\documentclass[10pt, twocolumn, letterpaper]{article}

\usepackage{cvpr}
\usepackage{times}
\usepackage{epsfig}
\usepackage{graphicx}
\usepackage{amsmath}
\usepackage{amssymb}
\usepackage{booktabs}
\usepackage{graphbox}
\usepackage{authblk}
\usepackage[utf8]{inputenc}

\usepackage[pagebackref=true,breaklinks=true,letterpaper=true,colorlinks,bookmarks=false]{hyperref}

\newcommand{\ours}{\texttt{ours}\xspace}
\newcommand{\tran}{Tran et al.\cite{tran2017regressing}\xspace}
\newcommand{\mofa}{\texttt{MoFA\cite{tewari17MoFA}}\xspace}

\cvprfinalcopy

\def\cvprPaperID{1349}

\ifcvprfinal\fi
\begin{document}

\title{Unsupervised Training for 3D Morphable Model Regression\vspace{-2ex}}
\makeatletter
\renewcommand\AB@affilsepx{\qquad \protect\Affilfont}
\makeatother
\author[1,2]{Kyle Genova}
\author[2]{Forrester Cole}
\author[2]{Aaron Maschinot}
\author[2]{Aaron Sarna}
\author[2]{Daniel Vlasic}
\author[2,3]{William T. Freeman}
\affil[1]{Princeton University}
\affil[2]{Google Research}
\affil[3]{MIT CSAIL}
\renewcommand*{\Authsep}{ }
\renewcommand*{\Authand}{ }
\renewcommand*{\Authands}{ }
\setlength{\affilsep}{0.5em}   % set the space between author and affiliation

\maketitle

\begin{abstract}

We present a method for training a regression network from image pixels to 3D morphable model coordinates using only unlabeled photographs. The training loss is based on features from a facial recognition network, computed on-the-fly by rendering the predicted faces with a differentiable renderer. To make training from features feasible and avoid network fooling effects, we introduce three objectives: a batch distribution loss that encourages the output distribution to match the distribution of the morphable model, a loopback loss that ensures the network can correctly reinterpret its own output, and a multi-view identity loss that compares the features of the predicted 3D face and the input photograph from multiple viewing angles. We train a regression network using these objectives, a set of unlabeled photographs, and the morphable model itself, and demonstrate state-of-the-art results.

\end{abstract}

\section{Introduction}

\begin{figure}
\begin{tabular}{ccc}
\includegraphics[width=1.0in]{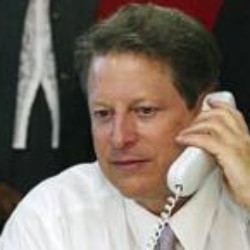} &
\includegraphics[width=1.0in]{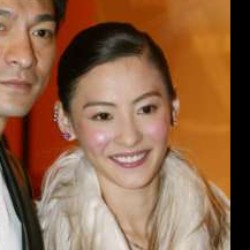} &
\includegraphics[width=1.0in]{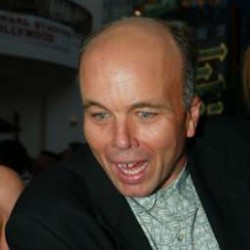} \\

\includegraphics[width=1.0in]{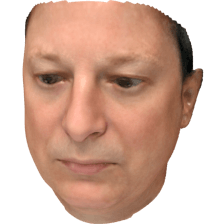} &
\includegraphics[width=1.0in]{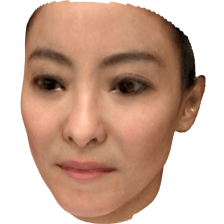} &
\includegraphics[width=1.0in]{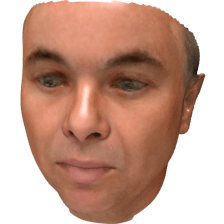}\\[0.5cm]

\includegraphics[width=1.0in]{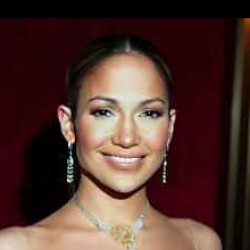} &
\includegraphics[width=1.0in]{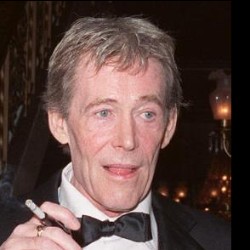} &
\includegraphics[width=1.0in]{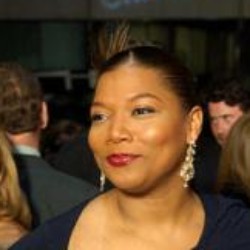}\\

\includegraphics[width=1.0in]{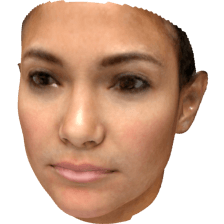} &
\includegraphics[width=1.0in]{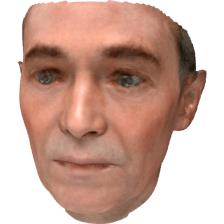} &
\includegraphics[width=1.0in]{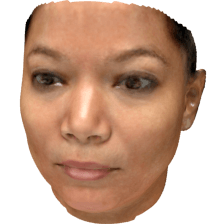}\\

\end{tabular}
\caption{Neutral 3D faces computed from input photographs using our regression network. We map features from a facial recognition network~\cite{7298682} into identity parameters for the Basel 2017 Morphable Face Model~\cite{basel2017}.}
\end{figure}

A 3D morphable face model (3DMM)~\cite{blanz1999morphable} provides a smooth, low-dimensional ``face space'' spanning the range of human appearance. 
Finding the coordinates of a person in this space from a single image of that person is a common task for applications such as 3D avatar creation, facial animation transfer, and video editing (e.g.~\cite{reanimating_cgf,Garrido:2016,thies2016face}).
The conventional approach is to search the space through inverse rendering, which generates a face that matches the photograph by optimizing shape, texture, pose, and lighting parameters~\cite{Levine:2009:SFR}. This approach requires a complex, non-linear optimization that can be difficult to solve in practice.

Recent work has demonstrated fast, robust fitting by regressing from image pixels to morphable model coordinates using a neural network~\cite{7785121,Richardson_2017_CVPR,tran2017regressing,tewari17MoFA}. The major issue with the regression approach is the lack of ground-truth 3D face data for training. Scans of face geometry and texture are difficult to acquire, both because of expense and privacy considerations. Previous approaches have explored synthesizing training pairs of image and morphable model coordinates in a preprocess~\cite{7785121, Richardson_2017_CVPR, tran2017regressing}, or training an image-to-image autoencoder with a fixed, morphable-model-based decoder and an image-based loss~\cite{tewari17MoFA}.  

This paper presents a method for training a regression network that removes both the need for supervised training data and the reliance on inverse rendering to reproduce image pixels. Instead, the network learns to minimize a loss based on the facial identity features produced by a face recognition network such as VGG-Face~\cite{Parkhi15} or Google's FaceNet~\cite{7298682}. These features are robust to pose, expression, lighting, and even non-photorealistic inputs. We exploit this invariance to apply a loss that matches the identity features between the input photograph and a synthetic rendering of the predicted face. The synthetic rendering need not have the same pose, expression, or lighting of the photograph, allowing our network to predict only shape and texture.

Simply optimizing for similarity between identity features, however, can teach the regression network to \emph{fool} the recognition network by producing faces that match closely in feature space but look unnatural. We alleviate the fooling problem by applying three novel losses: a \emph{batch distribution loss} to match the statistics of each training batch to the statistics of the morphable model, a \emph{loopback loss} to ensure the regression network can correctly reinterpret its own output, and a \emph{multi-view identity loss} that combines features from multiple, independent views of the predicted shape.

Using this scheme, we train a 3D shape and texture regression network using only a face recognition network, a morphable face model, and a dataset of unlabeled face images. We show that despite learning from unlabeled photographs, the 3D face results improve on the accuracy of previous work and are often recognizable as the original subjects.

\begin{figure*}
\includegraphics[width=2.09\columnwidth]{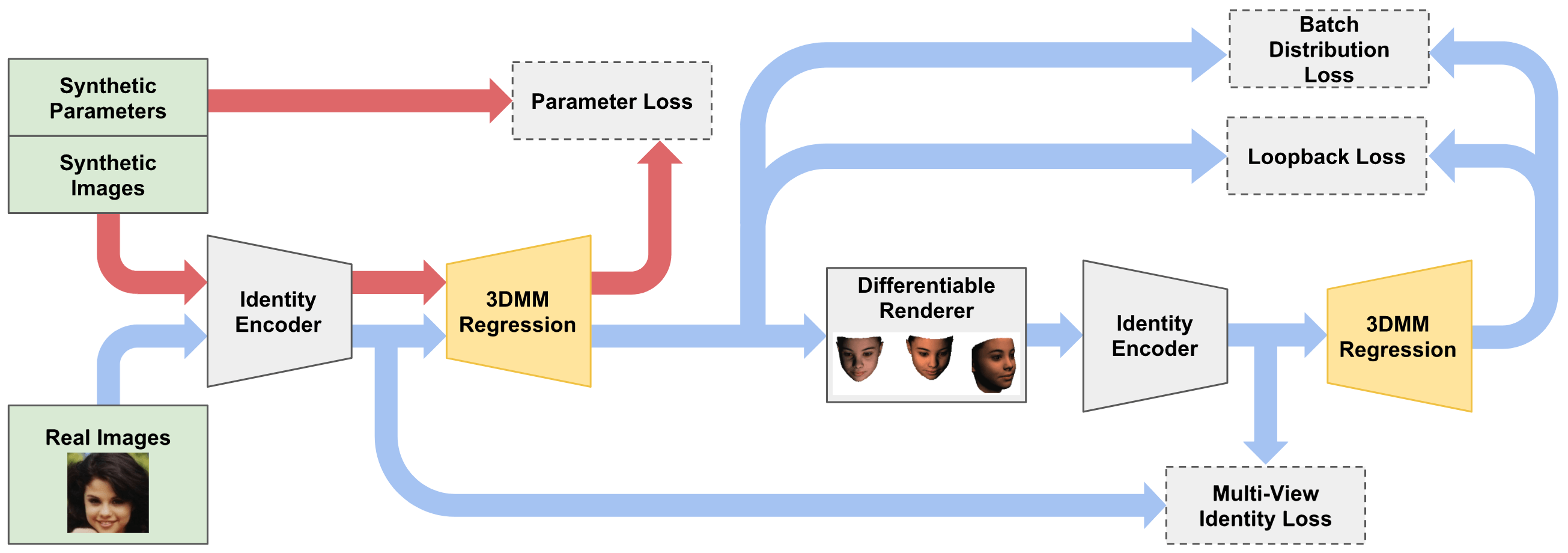}
\caption{End-to-end computation graph for unsupervised training of the 3DMM regression network. Training batches consist of combinations of real (blue) and synthetic (red) face images. Identity, loopback and batch distribution losses are applied to real images, while the 3DMM parameter loss is applied to synthetic images. The regression network (yellow) is shown in two places, but both correspond to the same instance during training. The identity encoder network is fixed during training.}
\label{fig:model_layout}
\end{figure*}

\section{Related Work}

\subsection{Morphable 3D Face Models}
Blanz and Vetter~\cite{blanz1999morphable} introduced the 3D morphable face model as an extension of the 2D active appearance model~\cite{cootes1998aam}. They demonstrated face reconstruction from a single image by iteratively fitting a linear combination of registered scans and pose, camera, and lighting parameters. They decomposed the geometry and texture of the face scans using PCA to produce separate, reduced-dimension geometry and texture spaces. Later work~\cite{basel2017} added more face scans and extended the model to include expressions as another separate space. We build directly off of this work by using the PCA weights as the output of our network. 

Convergence of iterative fitting is sensitive to the initial conditions and the complexity of the scene (i.e., lighting, expression, and pose). Subsequent work (\cite{Blanz:2003:FRB,Romdhani:2005:EST,thies2016face,Garrido:2016,Levine:2009:SFR} and others) has applied a range of techniques to improve the accuracy and stability of the fitting, producing very accurate results under good conditions. However, iterative approaches are still unreliable under general, in-the-wild, conditions, leading to the interest in regression-based approaches.

\subsection{Learning to Generate 3D Face Models}

Deep neural networks provide the ability to learn a regression from image pixels to 3D model parameters. The chief difficulty becomes how to collect enough training data to feed the network.  

One solution is to generate synthetic training data by drawing random samples from the morphable model and rendering the resulting faces~\cite{7785121, Richardson_2017_CVPR}. However, a network trained on purely synthetic data may perform poorly when faced with occlusions, unusual lighting, or ethnicities that are not well-represented by the morphable model. We include randomly generated, synthetic faces in each training batch to provide ground truth 3D coordinates, but train the network on real photographs at the same time.

Tran et al. \cite{tran2017regressing} address the lack of training data by using an iterative optimization to fit an expressionless model to a large number of photographs, and treat results where the optimization converged as ground truth. To generalize to faces with expression, identity labels and at least one neutral image are required, so the potential size of the training dataset is restricted. We also directly predict a neutral expression, but our unsupervised approach removes the need for an initial iterative fitting step.

An approach closely related to ours was recently proposed by Tewari, et al.~\cite{tewari17MoFA}, who train an autoencoder network on unlabeled photographs to predict shape, expression, texture, pose, and lighting simultaneously. The encoder is a regression network from images to morphable-model coordinates, and the decoder is a fixed, differentiable rendering layer that attempts to reproduce the input photograph. Like ours, this approach does not require supervised training pairs. However, since the training loss is based on individual image pixels, the network is vulnerable to confounding variation between related variables. For example, it cannot readily distinguish between dark skin tone and a dim lighting environment. Our approach exploits a pretrained face recognition network, which distinguishes such related variables by extracting and comparing features across the entire image. 

Other recent deep learning approaches predict depth maps \cite{sela2017unrestricted} or voxel grids \cite{jackson2017large}, trading off a compact and interpretable output mesh for more faithful reproductions of the input image. As for \cite{tewari17MoFA}, identity and expression are confounded in the output mesh. The result may be suitable for image processing tasks, such as relighting, at the expense of animation tasks such as rigging. 
%These works are complementary to ours in that they address a different set of use cases for a 3D face model.

%A second solution is to apply an offline, generative fitting method to produce training data\cite{tran2017regressing}. This approach allows fully-supervised training, but limits the training data to photographs for which the generative method succeeds. Our unsupervised approach removes the need for a generative model, and so avoids the issues with tuning and validating the generative results.

\subsection{Facial Identity Features}

Current face recognition networks achieve high accuracy over millions of identities \cite{kemelmacher2016megaface}. The networks operate by embedding images in a high-dimensional space, where images of the same person map to nearby points~\cite{7298682,Parkhi15,deepface}. Recent work~\cite{Cole2017SynthesizingNF,tewari17MoFA} has shown that this mapping is somewhat reversible, meaning the features can be used to produce a likeness of the original person. We build on this work and use FaceNet~\cite{7298682} to both produce input features for our regression network, and to verify that the output of the regression resembles the input photograph.

\section{Model}

We employ an encoder-decoder architecture that permits end-to-end unsupervised learning of 3D geometry and texture morphable model parameters (Fig.~\ref{fig:model_layout}). Our training framework utilizes a realistic, parameterized illumination model and differentiable renderer to form neutral-expression face images under varying pose and lighting conditions. We train our model on hybrid batches of real face images from VGG-Face~\cite{Parkhi15} and synthetic faces constructed from the Basel Face 3DMM~\cite{basel2017}.

The main strength and novelty of our approach lies in isolating our loss function to identity. By training the model to preserve identity through conditions of varying expression, pose, and illumination, we are able to avoid network fooling and achieve robust state-of-the-art recognizability in our predictions.

\subsection{Encoder}

We use FaceNet~\cite{7298682} for the network encoder, since its features have been shown to be effective for generating face images~\cite{Cole2017SynthesizingNF}. Other facial recognition networks such as VGG-Face~\cite{Parkhi15}, or even networks not focused on recognition, may work equally well.

The output of the encoder is the penultimate, 1024\nobreakdash-D \texttt{avgpool} layer of the ``NN2'' FaceNet architecture. We found the \texttt{avgpool} layer more effective than the final, 128-D \texttt{normalizing} layer as input to the decoder, but use the \texttt{normalizing} layer for our identity loss (Sec.~\ref{sec:id_loss}).  

\subsection{Decoder}

Given encoder outputs generated from a face image, our decoder generates parameters for the Basel Face Model 2017 3DMM~\cite{basel2017}.
The Basel 2017 model generates shape meshes $\mathbf{S}\equiv\left\{\mathbf{s}_i\in\mathbb{R}^3|1\le i\le N\right\}$ and texture meshes $\mathbf{T}\equiv\left\{\mathbf{t}_i\in\mathbb{R}^3|1\le i\le N\right\}$ with $N=53,149$ vertices.
\begin{equation}
    \begin{gathered}
        \mathbf{S}=\mathbf{S}(\mathbf{s},\mathbf{e})=\boldsymbol{\mu}_S+\mathbf{P}_{SS}\mathbf{W}_{SS}\mathbf{s}+\mathbf{P}_{SE}\mathbf{W}_{SE}\mathbf{e}\\
        \mathbf{T}=\mathbf{T}(\mathbf{t})=\boldsymbol{\mu}_T+\mathbf{P}_{T}\mathbf{W}_{T}\mathbf{t}
    \end{gathered}
    \label{eq:basel_meshes}
\end{equation}
Here, $\mathbf{s},\mathbf{t}\in\mathbb{R}^{199}$ and $\mathbf{e}\in\mathbb{R}^{100}$ are shape, texture, and expression parameterization vectors with standard normal distributions; $\boldsymbol{\mu}_S,\boldsymbol{\mu}_T\in\mathbb{R}^{3N}$ are the average face shape and texture; $\mathbf{P}_{SS},\mathbf{P}_{T}\in\mathbb{R}^{3N\times 199}$ and  $\mathbf{P}_{SE}\in\mathbb{R}^{3N\times 100}$ are linear PCA bases; and $\mathbf{W}_{SS},\mathbf{W}_{T}\in\mathbb{R}^{199\times 199}$ and $\mathbf{W}_{SE}\in\mathbb{R}^{100\times 100}$ are diagonal matrices containing the square roots of the corresponding PCA eigenvalues.

The decoder is trained to predict the $398$ parameters constituting the shape and texture vectors, $\mathbf{s}$ and $\mathbf{t}$, for a face. The expression vector $\mathbf{e}$ is not currently predicted and is set to zero. The decoder network consists of two $1024$-unit fully connected + ReLU layers followed by a $398$-unit regression layer. The weights were regularized towards zero. Deeper networks were considered, but they did not significantly improve performance and were prone to overfitting.

\subsubsection{Differentiable Renderer}

In contrast to previous approaches~\cite{Richardson_2017_CVPR, tewari17MoFA} that backpropagate loss through an image, we employ a general-purpose, differentiable rasterizer based on a deferred shading model. The rasterizer produces screen-space buffers containing triangle IDs and barycentric coordinates at each pixel. After rasterization, per-vertex attributes such as colors and normals are interpolated at the pixels using the barycentric coordinates and IDs. This approach allows rendering with full perspective and any lighting model that can be computed in screen-space, which prevents image quality from being a bottleneck to accurate training. The source code for the renderer is publicly available\footnote{\url{http://github.com/google/tf_mesh_renderer}}.

The rasterization derivatives are computed for the barycentric coordinates, but not the triangle IDs. We extend the definition of the derivative of barycentric coordinates with respect to vertex positions to include negative barycentric coordinates, which lie outside the border of a triangle. Including negative barycentric coordinates and omitting triangle IDs effectively treats the shape as locally planar, which is an acceptable approximation away from occlusion boundaries. Faces are largely smooth shapes with few occlusion boundaries, so this approximation is effective in our case, but it could pose problems if the primary source of loss is related to translation or occlusion.

\subsubsection{Illumination Model}
Because our differentiable renderer uses deferred shading, illumination is computed independently per-pixel with a set of interpolated vertex attribute buffers computed for each image. % TODO: Deferred shading is common, but maybe we could cite it? I had some trouble googling the source.
We use the Phong reflection model~\cite{Phong:1975:ICG} for shading. Because human faces exhibit specular highlights, Phong reflection allows for improved realism over purely diffuse approximations, such as those used in MoFA~\cite{tewari17MoFA}. It is both efficient to evaluate and differentiable.

To create appropriately even lighting, we randomly position two point light sources of varying intensity several meters from the face to be illuminated.
%Randomly perturbing light positions ensures that specular highlights on the rendered face model move appropriately.
We select a random color temperature for each training image from approximations of common indoor and outdoor light sources, and perturb the color to avoid overfitting. Finally, since the Basel Face model does not contain specular color information, we use a heuristic to define specular colors $K_s$ from the diffuse colors $K_d$ of the predicted model: $K_s := c - cK_d$ for some manually selected constant $c \in [0,1]$.

\subsection{Losses}

\begin{figure}
\begin{tabular}{ccc}
  Input & No Batch Dist. & Full Model \\
  \includegraphics[width=0.28\columnwidth]{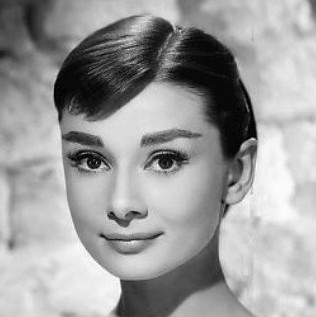} &
  \includegraphics[width=0.28\columnwidth]{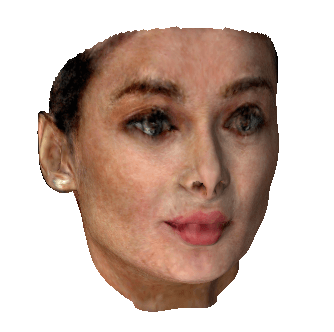} &
  \includegraphics[width=0.28\columnwidth]{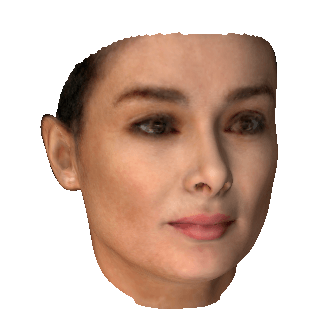} \\
  & No Loopback & \\
  \includegraphics[width=0.28\columnwidth]{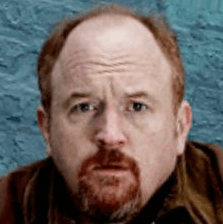} &
  \includegraphics[width=0.28\columnwidth]{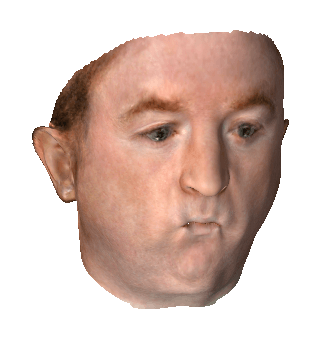} &
  \includegraphics[width=0.28\columnwidth]{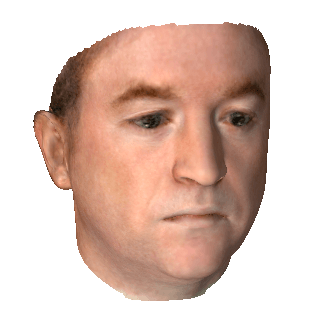} \\
  & No Multi-View & \\
  \includegraphics[width=0.28\columnwidth]{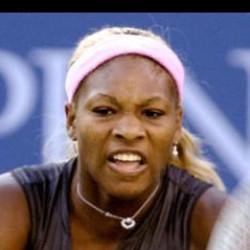} &
  \includegraphics[width=0.28\columnwidth]{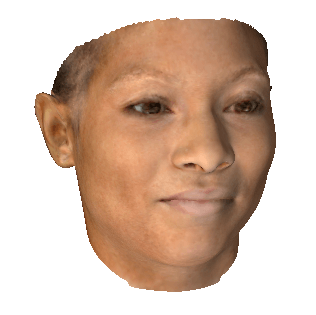} &
  \includegraphics[width=0.28\columnwidth]{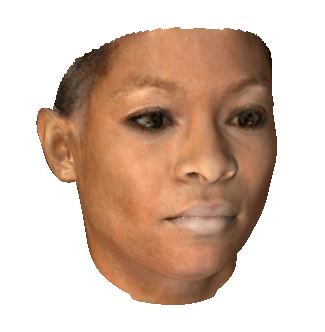} \\
\end{tabular}
\caption{Ablation test showing failures caused by removing individual losses. Batch distribution (top) keeps the results in the space of human faces, while loopback (middle) helps avoid exaggerated features. Multi-view identity (bottom) increases robustness to expression and lighting variation. Ablated result is computed by rendering a single frontal view for identity loss.}
\label{fig:ablation}
\end{figure}

We propose a novel loss function that focuses on facial identity, and ignores variations in facial expression, illumination, pose, occlusion, and resolution. This loss function is conceptually straightforward and enables unsupervised end-to-end training of our network. It combines four terms:
\begin{align}
    \begin{split}
        L=&L_{param}+L_{id}+\omega_{batch}L_{batch}+\omega_{loop}L_{loop}
    \end{split}
    \label{loss_func}
\end{align}
Here, $L_{param}$ imposes 3D shape and texture similarity for the synthetic images, $L_{id}$ imposes identity preservation on the real images in a batch, $L_{batchdistr}$ regularizes the predicted parameter distributions within a batch to the distribution of the morphable model, and $L_{loopback}$ ensures the network can correctly interpret its own output. The effects of removing the batch distribution, loopback, and limiting the identity loss to a single view are shown in Figure~\ref{fig:ablation}. We use $\omega_{batch}=10.0$ and $\omega_{loop}=0.07$ for our results. 

Training proceeds in two stages. First, the model is trained solely on batches of synthetic faces generated by randomly sampling for shape, texture, pose, and illumination parameters. This stage performs only a partial training of the model: since shape and texture parameters are sampled independently in this stage, the model is restricted from learning correlations between them. Second, the partially-trained model is trained to convergence on batches consisting of a combination of real face images from the VGG-Face dataset~\cite{Parkhi15} and synthetic faces. Synthetic faces are subject to only the $L_{param}$ loss, while real face images are subject to all losses except $L_{param}$. 
%components of the loss function in~\eqref{loss_func} except for $L_{param}$.

\subsubsection{Parameter Loss}
\label{sec:param_loss}

For synthetic faces, the true shape and texture parameters are known, so we use independent Euclidean losses between the randomly generated true synthetic parameter vectors, $\bar{\mathbf{s}}_b$ and $\bar{\mathbf{t}}_b$, and the predicted ones, $\mathbf{s}_b$ and $\mathbf{t}_b$, in a batch.
\begin{equation}
    L_{param}=\omega_{s}\sum_{b}\left|\mathbf{s}_b-\bar{\mathbf{s}}_b\right|^2+\omega_{t}\sum_{b}\left|\mathbf{t}_b-\bar{\mathbf{t}}_b\right|^2
\end{equation}
where $\omega_{s}$ and $\omega_{t}$ control the relative contribution of the shape and texture losses. Due to different units, we set $\omega_{s}=0.4$ and $\omega_{t}=0.002$.

\subsubsection{Identity Loss}
\label{sec:id_loss}

Robust prediction of recognizable meshes can be facilitated with a loss that derives from a facial recognition network. We used FaceNet~\cite{7298682}, though the identity-preserving loss generalizes to other networks such as VGG-Face~\cite{Parkhi15}. The final FaceNet \texttt{normalizing} layer is a 128-D unit vector such that, regardless of expression, pose, or illumination, same-identity inputs map closer to each other on the hypersphere than different-identity ones. For our identity loss $L_{id}$, we define similarity of two faces as the cosine score of their respective output unit vectors, $\boldsymbol{\gamma}_1$ and $\boldsymbol{\gamma}_2$:
\begin{equation}
    L_{id}\left(\boldsymbol{\gamma}_1,\boldsymbol{\gamma}_2\right)=\boldsymbol{\gamma}_1\cdot\boldsymbol{\gamma}_2
    \label{facenet_sim}
\end{equation}
To use this loss in an unsupervised manner on real faces, we calculate the cosine score between a face image and the image resulting from passing the decoder outputs into the differentiable renderer with random pose and illumination.

Identity prediction can be further enhanced by using multiple poses for each face. Multiple poses decrease the presence of occluded regions of the mesh. Additionally, since each pose provides a backpropagation path to the mesh vertices, the model trains in a more robust manner than if only a single pose is used. We use three randomly determined poses for each real face.

\subsubsection{Batch Distribution Loss}

Applying the identity loss alone allows training to introduce biases into the decoder outputs that change their distribution from the zero-mean standard normal distribution assumption made by the Basel Face Model. These changes are likely due to domain transfer effects between the real images and those rendered from the decoder outputs. Initially, we attempted to compensate for these effects by adding a shallow network to transform the model-rendered encoder outputs prior to calculating the identity loss. While this approach did increase overall recognizability in the model, it also introduced unrealistic artifacts into the model outputs.

Instead, we opted to regularize each batch~\cite{Salimans:2016:WN,karras2018progressive} to directly constrain the lowest two moments of the shape and texture parameter distributions to match those of a zero-mean standard normal distribution. This loss, which is applied at a batch level, combines four terms:
\begin{equation}
    \begin{gathered}
        L_{batchdistr}=L_{\mu_S}+L_{\sigma_S}+L_{\mu_T}+L_{\sigma_T}
    \label{batch_loss_func}
    \end{gathered}
\end{equation}
Here, $L_{\mu_S}$ and $L_{\mu_T}$ regularize the batch shape and texture parameters to have zero mean, and $L_{\sigma_S}$ and $L_{\sigma_T}$ regularize them to have unit variance.
\begin{equation}
    \begin{gathered}
        L_{\mu_S}=\left|Mean_b\left[\mathbf{s}_b\right]-\mathbf{0}_{199}\right|^2\\
        L_{\sigma_S}=\left|\ \,Var_b\left[\mathbf{s}_b\right]\,\,\,-\mathbf{1}_{199}\right|^2 \\
        L_{\mu_T}=\left|Mean_b\left[\mathbf{t}_b\right]-\mathbf{0}_{199}\right|^2 \\
        L_{\sigma_T}=\left|\ \,Var_b\left[\mathbf{t}_b\right]\,\,\,-\mathbf{1}_{199}\right|^2
    \end{gathered}
\end{equation}

\subsubsection{Loopback Loss}

A limitation of using real face images for unsupervised training is that the true shape and texture parameters for the faces are unknown. If they were known, then the more direct lower-level parameter loss in Sec.\,\ref{sec:param_loss} could be directly imposed instead of the identity loss in Sec.\,\ref{sec:id_loss}.

A close approximation to this lower-level loss for real images can be achieved using a ``loopback'' loss (Fig.~\ref{fig:model_layout}). The nature of this loss lies in generalizing the model near the regions for which real face image data exists. Similar techniques have proven to be successful in generalizing model learning for image applications\cite{Nair:2008:ALI,Li:2017:MSA}.

To compute the loopback loss at any training step, the current-state decoder outputs for a batch of real face images are extracted and used to generate synthetic faces rendered in random poses and illuminations. The synthetic faces are then passed back through the encoder and decoder again, and the parameter loss in Sec.\,\ref{sec:param_loss} is imposed between the resulting parameters and those first output by the decoder.

As shown in Fig.~\ref{fig:model_layout}, two loopback loss backpropagation paths to the decoder exist. The effects of each are complementary: the synthetic face parameter path generalizes the decoder in the region near that of real face parameters, and the real image channel regularizes the decoder away from generating unrealistic faces. Additionally, the two paths encourage the regression network to match its responses for real and synthetic versions of the same face.

\section{Experiments}

\begin{figure*}
%\begin{tabular}{cccccccc}
\begin{tabular}{cc@{\hspace{3mm}}c@{\hspace{3mm}}c@{\hspace{3mm}}c@{\hspace{3mm}}c@{\hspace{3mm}}c@{\hspace{3mm}}c}
% Inputs
\vspace{.4em}
\hspace{-0.7cm}{Input} &
\includegraphics[width=0.80in,align=c]{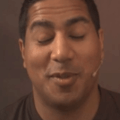} &
\includegraphics[width=0.80in,align=c]{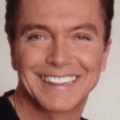} &
\includegraphics[width=0.80in,align=c]{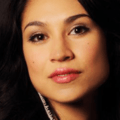} &
\includegraphics[width=0.80in,align=c]{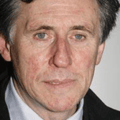} &
\includegraphics[width=0.80in,align=c]{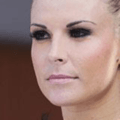} &
\includegraphics[width=0.80in,align=c]{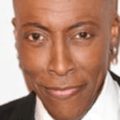} &
\includegraphics[width=0.80in,align=c]{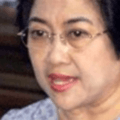}\\

% Us
\vspace{.4em}
% Revision: nov8, from large set.
\hspace{-0.7cm}{Ours} &
\includegraphics[width=0.80in,align=c]{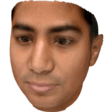} &
\includegraphics[width=0.80in,align=c]{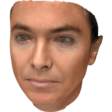} &
\includegraphics[width=0.80in,align=c]{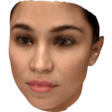} &
\includegraphics[width=0.80in,align=c]{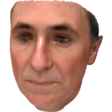} &
\includegraphics[width=0.80in,align=c]{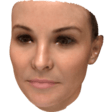} &
\includegraphics[width=0.80in,align=c]{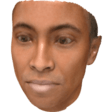} &
\includegraphics[width=0.80in,align=c]{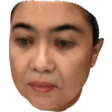} \\

% Tran
\vspace{.4em}
\hspace{-0.7cm}{Tran\cite{tran2017regressing}} &
%\cite{tran2017regressing} &
\includegraphics[width=0.80in,align=c]{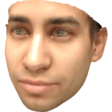} &
\includegraphics[width=0.80in,align=c]{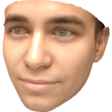} &
\includegraphics[width=0.80in,align=c]{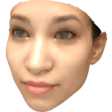} &
\includegraphics[width=0.80in,align=c]{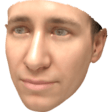} &
\includegraphics[width=0.80in,align=c]{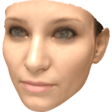} &
\includegraphics[width=0.80in,align=c]{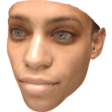} &
\includegraphics[width=0.80in,align=c]{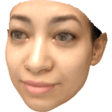}\\

% MoFA
\vspace{.4em}
\hspace{-0.7cm}{MoFA\cite{tewari17MoFA}} &
\includegraphics[width=0.80in,align=c]{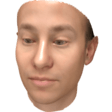} &
\includegraphics[width=0.80in,align=c]{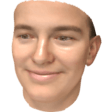} &
\includegraphics[width=0.80in,align=c]{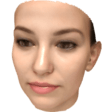} &
\includegraphics[width=0.80in,align=c]{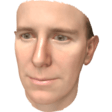} &
\includegraphics[width=0.80in,align=c]{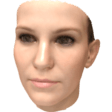} &
\includegraphics[width=0.80in,align=c]{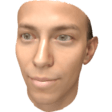} &
\includegraphics[width=0.80in,align=c]{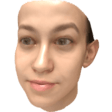}\\
% MoFA
\vspace{.4em}
\hspace{-0.7cm}{MoFA+Exp} &
\includegraphics[width=0.80in,align=c]{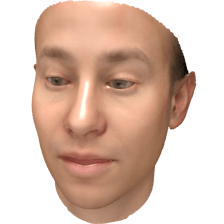} &
\includegraphics[width=0.80in,align=c]{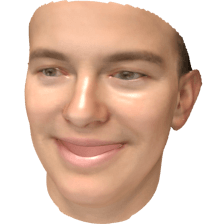} &
\includegraphics[width=0.80in,align=c]{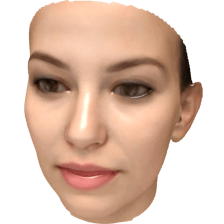} &
\includegraphics[width=0.80in,align=c]{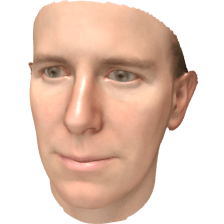} &
\includegraphics[width=0.80in,align=c]{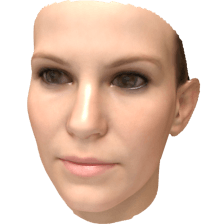} &
\includegraphics[width=0.80in,align=c]{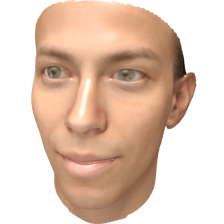} &
\includegraphics[width=0.80in,align=c]{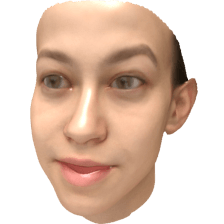}\\
% Sela
\vspace{.4em}
\hspace{-0.7cm}{Sela\cite{sela2017unrestricted}} &
\includegraphics[width=0.80in,align=c]{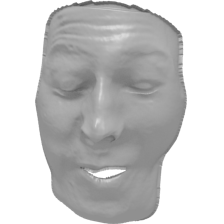} &
\includegraphics[width=0.80in,align=c]{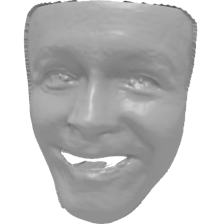} &
\includegraphics[width=0.80in,align=c]{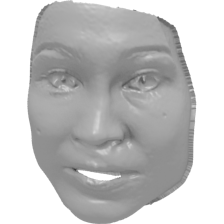} &
\includegraphics[width=0.80in,align=c]{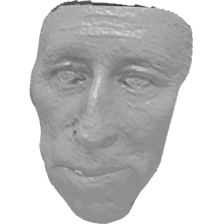} &
\includegraphics[width=0.80in,align=c]{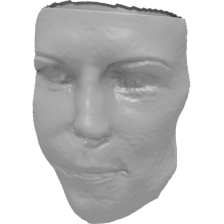} &
\includegraphics[width=0.80in,align=c]{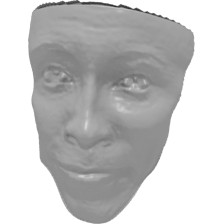} &
\includegraphics[width=0.80in,align=c]{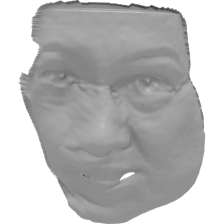}\\
\end{tabular}
\caption{Results on the MoFA-Test dataset. Our method shows improved likeness and color fidelity over competing methods, especially in the shape of the eyes, eyebrows, and nose. Note that MoFA~\cite{tewari17MoFA}  solves for pose, expression, lighting, and identity, so is shown both with (row 5) and without (row 4) expression. The unstructured method of Sela, et al.~\cite{sela2017unrestricted} produces only geometry, so is shown without color.}
\label{fig:qualitative}
\end{figure*}

We first show and discuss the qualitative improvements of our method compared with other morphable model regression approaches (Sec.~\ref{sec:qualitative_comparison}). We then evaluate our method quantitatively by comparing reconstruction error against scanned 3D face geometry (Sec.~\ref{sec:micc}) and features produced by VGG-Face, which was not used for training (Sec.~\ref{sec:face_recognition_results} and \ref{sec:face_clustering}). We also show qualitative results on corrupted and non-photorealistic inputs (Sec.~\ref{sec:stress_tests}).

\subsection{Qualitative Comparison}
\label{sec:qualitative_comparison}

Figure~\ref{fig:qualitative} compares our results with the methods of of Tran, et al.~\cite{tran2017regressing}, Tewari, et al.~\cite{tewari17MoFA} (MoFA), and Sela, et al.~\cite{sela2017unrestricted} on 7 images from an 84-image test set developed as part of MoFA. An extended evaluation is available in the supplemental material. Our method improves on the likenesses of previous approaches, especially in features relevant to facial recognition such as the eyebrow texture and nose shape. 

Our method also predicts coloration and skin tone more faithfully. This improvement is likely a consequence of our batch distribution loss, which allows individual faces to vary from the mean of the Basel model (light skin tone), so long as the faces match the mean \emph{in aggregate}. Previous methods, by contrast, regularize \emph{each face} towards the mean of the model's distribution, tending to produce light skin tone overall.

The MoFA approach also sometimes confounds identity and expression (Fig.~\ref{fig:qualitative}, second column), and skin tone and lighting (Fig.~\ref{fig:qualitative}, first and sixth columns). Our method and Tran et al.~\cite{tran2017regressing} are more resistant to confounding variables. The unstructured method of Sela et al.~\cite{sela2017unrestricted} does not separate identity and expression, predicts only shape, and is less robust than the model-based methods.

\subsection{Neutral Pose Reconstruction on MICC}
\label{sec:micc}

We quantitatively evaluate the ground-truth accuracy of our models on the MICC Florence 3D Faces dataset~\cite{Bagdanov:2011:FHF:2072572.2072597} (MICC) in Table~\ref{table:micc}. This dataset contains the ground truth scans of 53 Caucasian subjects in a neutral expression. Accompanying the scans are three observation videos for each subject, in conditions of increasing difficulty: `cooperative', `indoor', and `outdoor.' We run the methods on each frame of the videos, and average the results over each video to get a single reconstruction. The results of Tran et al.~\cite{tran2017regressing} are averaged over the mesh, as in \cite{tran2017regressing}. We instead average our encoder embeddings before making a single reconstruction.

% Bl10 Results
\begin{table}[ht]
%\centering
\begin{tabular}{lc@{\hspace{0.5mm}}c|c@{\hspace{0.5mm}}c|c@{\hspace{0.5mm}}c}
\toprule
& \multicolumn{2}{c|}{Cooperative} & \multicolumn{2}{c|}{Indoor} & \multicolumn{2}{c}{Outdoor}\\
Method & Mean & Std. & Mean & Std. & Mean & Std.\\
\cmidrule(lr){2-3}
\cmidrule(lr){4-5}
\cmidrule(lr){6-7}
% Our Results: (Revision: bl10, without 6 or 22)
\tran & 1.93 & 0.27 & 2.02 & 0.25 & 1.86 & 0.23\\
\ours & \textbf{1.50} & \textbf{0.13} & \textbf{1.50} & \textbf{0.11} & \textbf{1.48} & \textbf{0.11}\\
\bottomrule
\end{tabular}
\vspace{0.5em}
\caption{Mean Error on MICC Dataset using point-to-plane distance after ICP alignment of video-averaged outputs with isotropic scale estimation. Our errors lower on average and in variance, both within individual subjects and as conditions change.}
\label{table:micc}
\end{table}

% Batch-reg Results
% \begin{table*}
% \centering
% \begin{tabular}{lcc|cc|cc}
% \toprule
% & \multicolumn{2}{c|}{Cooperative} & \multicolumn{2}{c|}{Indoor} & \multicolumn{2}{c}{Outdoor}\\
% Method & Mean & Std. Dev. (10x) & Mean & Std. Dev. (10x) & Mean & Std. Dev. (10x)\\
% \cmidrule(lr){2-3}
% \cmidrule(lr){4-5}
% \cmidrule(lr){6-7}
% \tran & 1.94 & 2.85 & 2.02 & 2.38 & 1.85 & 2.38\\
% % Our Results: (Revision: batch-reg, without 6 or 22)
% \ours &\textbf{1.37} &\textbf{0.88}& \textbf{1.36} & \textbf{0.95} & \textbf{1.34} & \textbf{0.87}\\
% \bottomrule
% \end{tabular}
% \vspace{0.5em}
% \caption{Mean Error on MICC Dataset using point-to-plane distance after ICP alignment of video-averaged outputs with isotropic scale estimation. Our results are between 27.6-32.7\% better on an absolute scale, and are much more consistent, both within individual subjects and as conditions change.}
% \label{table:micc}
% \end{table*}

% \begin{table}
% \centering
% \begin{tabular}{lccc}
% \toprule
% Method & Coop & Indoor & Outdoor\\
% \midrule
% \tran & 1.94 & 2.02 & 1.85\\
% % Our Results: (Revision: batch-reg, without 6 or 22)
% \ours & \textbf{1.37} & \textbf{1.36} & \textbf{1.34}\\
% \bottomrule
% \end{tabular}
% \vspace{0.5em}
% \caption{Mean Error on MICC Dataset using point-to-plane distance after ICP alignment of video-averaged outputs with isotropic scale estimation. Our results are between 27.6-32.7\% better on an absolute scale, and are much more consistent as conditions change.}
% \label{table:micc}
% \end{table}

To evaluate our predictions, we crop the ground truth scan to 95mm around the tip of the nose as in \cite{tran2017regressing}, and run ICP with isotropic scale to find an alignment. We solve for isotropic scale because we do not assume the methods predict absolute scale, and a small misalignment in scale can have a large effect on error. Table~\ref{table:micc} shows the symmetric point-to-plane distance in millimeters within the ICP-determined region of intersection, rather than point-to-point distances, as the methods and ground truth have different vertex densities. Our results indicate that we have improved absolute error to the ground truth by 20-25\%, and our results are more consistent from person to person, with less than half the standard deviation when compared to \cite{tran2017regressing}. We are also more stable across changing environments, with similar results for all three test sets.

\subsection{Face Recognition Results}
\label{sec:face_recognition_results}

\begin{figure}
\centering
\hspace{-0.15cm}{
\includegraphics[width=3.3in]{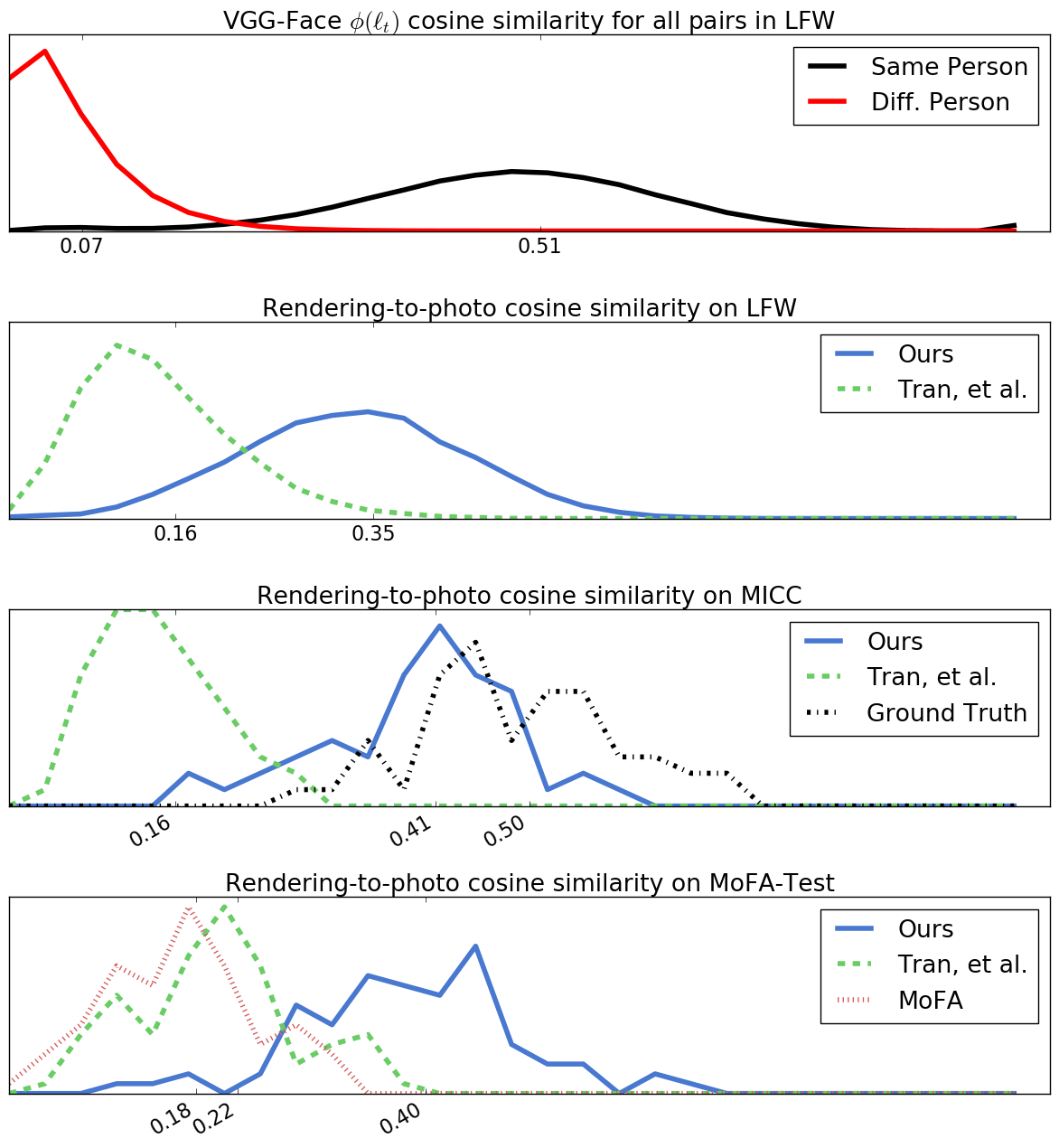}}
\vspace{-1.2em}
\caption{Distributions of cosine similarity between VGG-Face $\phi(\ell_t)$ layers for LFW, MICC, and MoFA-Test. Top: the similarity scores for all pairs of photos in LFW, divided into same and different person distributions. Below: similarity scores of our method, Tran, et al.~\cite{tran2017regressing}, and MoFA~\cite{tewari17MoFA} for photos and their corresponding 3D renderings on LFW, MICC, MoFA-Test. Mean values for each distribution are shown below. Camera and lighting parameters were fixed for all renderings. 
%Ground truth for MICC compares the photo with a rendering of the 3D face scan.
%In both cases, our method's distribution is closer to the same-person distribution than previous work. 
}
\label{fig:histograms}
\vspace{-1.0em}
\end{figure}

In order to quantitatively evaluate the likeness of our reconstructions, we use the VGG-Face~\cite{Parkhi15} recognition network's activations as a measure of similarity. VGG-Face was chosen because FaceNet appears in our training loss, making it unsuitable as an evaluation metric. For each face in our evaluation datasets, we compute the cosine similarity of the $\phi(\ell_t)$ layers of VGG-Face between the input image and a rendering of our output geometry, as described in \cite{Parkhi15}. 

The similarity distributions for Labeled Faces in the Wild~\cite{LFWTech} (LFW), MICC, and MoFA-Test are shown in Figure~\ref{fig:histograms}. 
The similarity between all pairs of photographs in the LFW dataset, separated into same-person and different-person distributions, is shown for comparison in Fig.~\ref{fig:histograms}, top. Our method achieves an average similarity between rendering and photo of 0.403 on MoFA test (the dataset for which results for all methods are available). By comparison, 22.7\% of pairs of photos of the same person in LFW have a score below 0.403, and only 0.04\% of pairs of photos of different people have a score above 0.403. 

For additional validation, Table~\ref{table:vgg_emd} shows the Earth Mover's distance~\cite{pele2009} between the all-pairs LFW distributions and the results of each method. Our method's results are closer to the same-person distribution than the different-person distribution in all cases, while the other methods results' are closer to the different-person distribution. We conclude that ours is the first method that generates neutral-pose, 3D faces with recognizability approaching a photo.

The scores of the ground-truth 3D face scans from MICC and their input photographs provide a ceiling for similarity scores. Notably, the distance between the GT distribution and the same-person LFW distribution is very low, with almost the same mean ($0.51$ vs $0.50$), indicating the VGG-Face network has little trouble bridging the domain gap between photograph and rendering, and that our method does not yet reach the ground-truth baseline.

\begin{table}
\centering
\begin{tabular}{lp{0.55cm}p{0.55cm}|p{0.55cm}p{0.55cm}|p{0.55cm}p{0.55cm}}
\toprule
& \multicolumn{2}{c|}{LFW} &
\multicolumn{2}{c|}{MICC} & \multicolumn{2}{c}{MoFA-T}\\
Method & Same & Diff. & Same & Diff.& Same & Diff.\\
\cmidrule(lr){2-3}
\cmidrule(lr){4-5}
\cmidrule(lr){6-7}
\mofa & $-$&$-$ & $-$&$-$ & 0.30 & 0.11\\
\tran & 0.32 & 0.09 & 0.32 & 0.09 & 0.27 & 0.14\\
\ours & \textbf{0.14} &\textbf{0.26} & 0.09 & 0.32& \textbf{0.09} & \textbf{0.32}\\
GT & $-$ & $-$ & \textbf{0.03} & \textbf{0.41} & $-$ & $-$\\
\bottomrule
\end{tabular}
\vspace{0.5em}
\caption{Earth mover's distance between distributions of VGG-Face $\phi(\ell_t)$ similarity and distributions of same and different identities on LFW. A low distance for ``Same'' means the similarity scores between a photo and its associated 3D rendering are close to the scores of same identity photos in LFW, while a low distance for ``Diff.'' means the scores are close to the scores of different identity photos.
}
\label{table:vgg_emd}
\end{table}

\subsection{Face Clustering}
\label{sec:face_clustering}

\begin{table}
\centering
\begin{tabular}{lcc|cc}
\toprule
& \multicolumn{2}{c|}{MoFA-Test} & \multicolumn{2}{c}{LFW}\\
Method & Top-1 & Top-5 & Top-1 & Top-5\\
\cmidrule(lr){2-3}
\cmidrule(lr){4-5}
\texttt{random} & 0.01 & 0.06 & 0.0002 & 0.001\\ %2.3089e-04
\mofa & 0.19 & 0.54 & $-$&$-$\\
\tran & 0.25 & 0.62 & 0.001 & 0.002\\ %Tran is 0.0011 and 0.0021
\ours & \textbf{0.87} &\textbf{0.96} & \textbf{0.16} & \textbf{0.51}\\
\bottomrule
\end{tabular}
\vspace{0.5em}
\caption{Identity Clustering Recall using VGG-Face distances on MoFA-Test and LFW. 
%100 photographs from LFW with no identity duplication were randomly selected, and both methods run on them. 
%Methods were run on the images of each dataset. Then, each generated mesh was rerendered, and 
Given a rendered mesh, the task is to recover the unknown source identity by looking up the nearest neighbor photographs according to VGG-Face $\phi(\ell_t)$ cosine similarity. Top-1 and Top-5 show the fractions for which a photograph of the correct identity was recalled as the nearest neighbor, or in the nearest 5, respectively. Performance is higher for MoFA-Test because it contains 84 images and 78 identities, while the LFW set contains 12,993 images and 5,749 identities.}
\label{table:face_clustering}
%\vspace{-1.0em}
\end{table}

To establish that our reconstructions are recognizable, we perform a clustering task to recover the identities of our generated meshes. For each of LFW and MoFA-Test, we run our method on all faces in the dataset, and render the output geometry as shown in the figures in this paper. For each rendering, we find the nearest neighbors according to the VGG-Face $\phi(\ell_t)$ distance. Table \ref{table:face_clustering} shows the fraction of meshes that recall a photo of the source identity as the nearest neighbor, and within the top 5 nearest neighbors. 

On MoFA-Test, which has 84 images and 78 identities, we achieve a Top-1 recall of 87\%, compared to 25\% for Tran et al. and 19\% for MoFA. On the larger LFW dataset, which contains over 5,000 identities in 13,000 photographs, we still achieve a Top-5 recall of 51\%.
We conclude our approach generates recognizable 3D morphable models, even in test sets with thousands of candidate identities.

\begin{figure}
\begin{tabular}{c@{\hspace{2mm}}c@{\hspace{2mm}}c@{\hspace{2mm}}c@{\hspace{2mm}}c@{\hspace{2mm}}}
% Inputs
% Revision: bl10
\hspace{-0.3cm}
\includegraphics[width=0.59in]{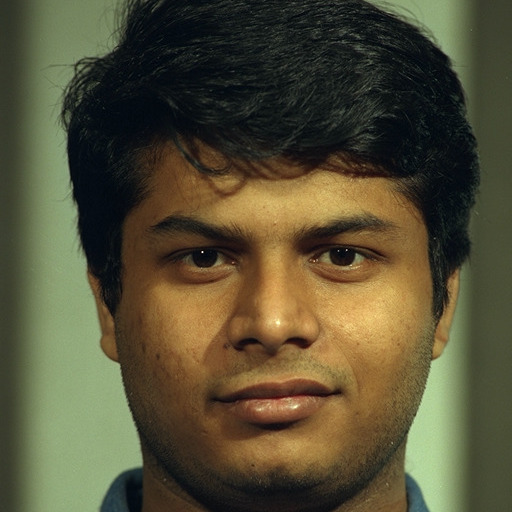} &
\includegraphics[width=0.59in]{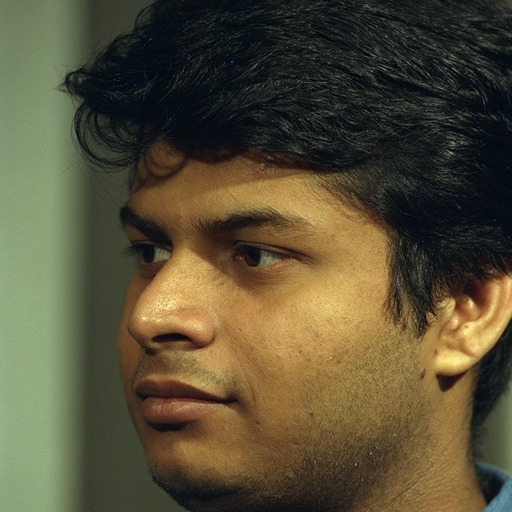} &
\includegraphics[width=0.59in]{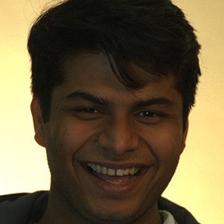} &
\includegraphics[width=0.59in]{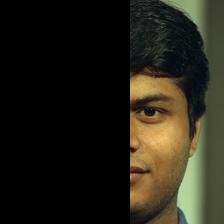} &
\includegraphics[width=0.59in]{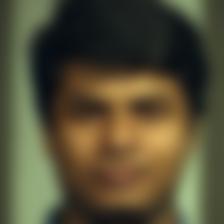}\\
% Us
\hspace{-0.3cm}
\includegraphics[width=0.59in]{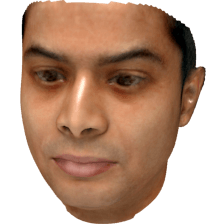} &
\includegraphics[width=0.59in]{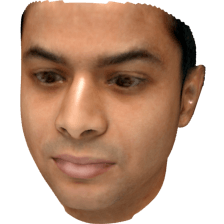} &
\includegraphics[width=0.59in]{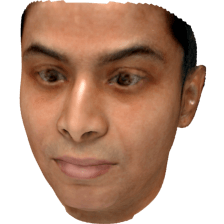} &
\includegraphics[width=0.59in]{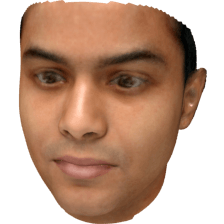} &
\includegraphics[width=0.59in]{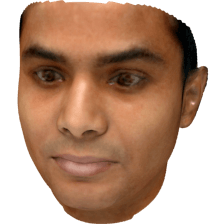}\\

\end{tabular}
\caption{FERET dataset~\cite{feret} stress test. The regression network is robust to changes in pose, lighting, expression, occlusion, and blur. See supplemental material for additional results.}
\label{fig:stress_test}
\vspace{-2.0mm}
\end{figure}

\subsection{Reconstruction from Challenging Images}
\label{sec:stress_tests}

Our regression network uses a facial identity feature vector as input, yielding results robust to changes in pose, expression, lighting, occlusion, and resolution, while remaining sensitive to changes in identity. Figure \ref{fig:stress_test} qualitatively demonstrates this robustness by varying conditions for a single subject and displaying consistent output.

\begin{figure}
\hspace{-1.2em}{
\begin{tabular}{ccc}
% Inputs
\includegraphics[width = 1.0in]{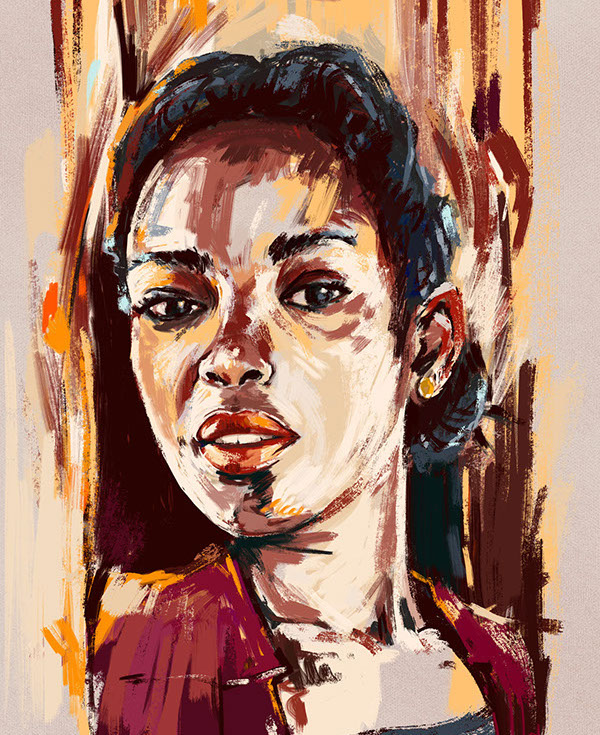} &
\includegraphics[width = 1.0in]{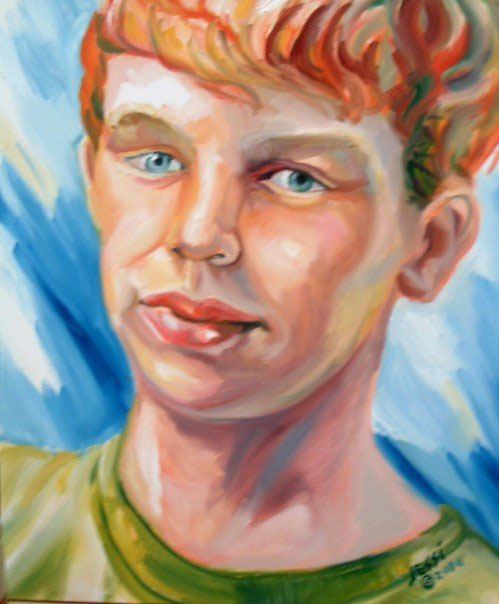} &
\includegraphics[width = 1.0in]{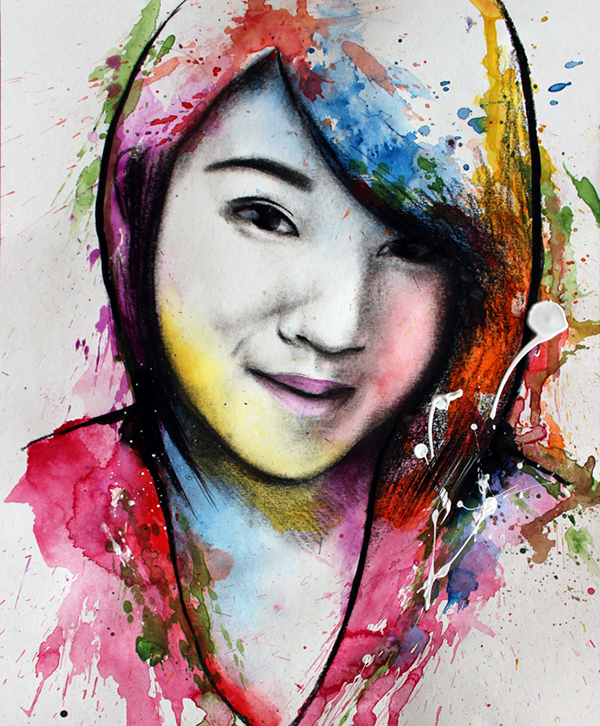}\\
% Us: Revision bl10
\includegraphics[width = 1.0in]{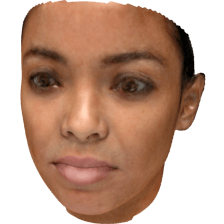} &
\includegraphics[width = 1.0in]{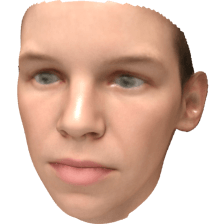} &
\includegraphics[width = 1.0in]{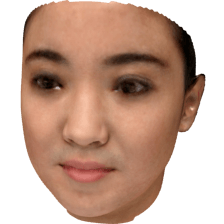}\\
\end{tabular}
\caption{Art from the BAM dataset~\cite{Wilber_2017_ICCV}. Because the inputs to our regression network are high-level identity features, the results are robust to stylized details at the pixel level. }
\label{fig:art}
}
\end{figure}

Additionally, Figure~\ref{fig:art} shows that our network can reconstruct plausible likenesses from non-photorealistic artwork, in cases where a fitting approach based on inverse rendering would have difficulty. This result is possible because of the invariance of the identity features to unrealistic pixel-level information, and because our unsupervised loss focuses on aspects of reconstruction that are important for recognition.

% \subsection{Fitting Pose and Expression}
% \label{sec:landmarks}

% \input{fig_fitting}

% Our system reconstructs shape and texture of faces, and ignores aspects such as pose, expression, and lighting. Those components are needed to exactly match the reconstruction to the source image, and our neutral face output is an excellent starting point to find them. Figure~\ref{fig:fitting} shows results of gradient descent that starts with our output and fits the pose and expression by minimizing the distances of landmarks on our mesh and the image (we used the 68 landmark configuration from the Multi-PIE database \cite{gross2008mpie}).

\section{Discussion and Future Work}

We have shown it is possible to train a regression network from images to neutral, expressionless 3D morphable model coordinates using only unlabeled photographs and improve on the accuracy of supervised methods. Our results  approach the face recognition similarity scores of real photographs and exceed the scores of other regression approaches by a large margin. Because of the accuracy of the approach, the predicted face can be directly used for face-tracking based on landmarks. 

This paper focuses on learning an expressionless face, which is suitable for creating VR avatars or landmark-based tracking. In future work, we hope to extend the approach to predict pose, expression, and lighting, similar to Tewari, et al.~\cite{tewari17MoFA}. Predicting these factors while avoiding their confounding effects should be possible by adding an inverse rendering stage to our decoder while maintaining the neutral-pose losses we currently apply.

The method produces generally superior results for young adults and Caucasian ethnicities. The differences could be due to limited representation in the scans used to produce the morphable model, bias in the features extracted from the face recognition network, or limited representation in the VGG-Face dataset we use for training. In future work, we hope to improve the performance of the method on a diverse range of ages and ethnicities. 

{\small
\bibliographystyle{ieee}
\bibliography{egbib}
}

\clearpage
\appendix
\section{Appendix}
% \begin{figure}

% \begin{tabular}{ccc}
% % Our revision: bl10
% %Input & This Paper\\%& Tran et al.\\
% \subfloat{\includegraphics[width=0.3\columnwidth]{lighting/input/cropped/00070_940307_hl.jpg}} &
% \subfloat{\includegraphics[width=0.3\columnwidth]{lighting/input/cropped/00070_940422_hl.jpg}} &
% \subfloat{\includegraphics[width=0.3\columnwidth]{lighting/input/cropped/00070_941121_ql.jpg}}\\

% \subfloat{\includegraphics[width=0.3\columnwidth]{lighting/us/00070_940307_hl.png}} &
% \subfloat{\includegraphics[width=0.3\columnwidth]{lighting/us/00070_940422_hl.png}} &
% \subfloat{\includegraphics[width=0.3\columnwidth]{lighting/us/00070_941121_ql.png}}\\

% \end{tabular}
% \caption{Stability under variable lighting conditions for a subject from the FERET~\cite{feret} dataset.}
% \label{fig:lighting}
% \end{figure}

% Vertical figure.

\begin{figure}[h]
\begin{tabular}{cc}
% Our revision: bl10
\includegraphics[width=0.45\columnwidth]{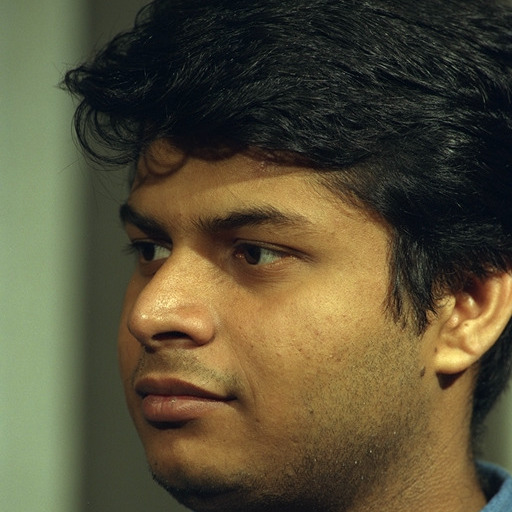} &
\includegraphics[width=0.45\columnwidth]{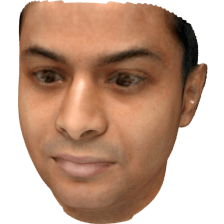}\\

\includegraphics[width=0.45\columnwidth]{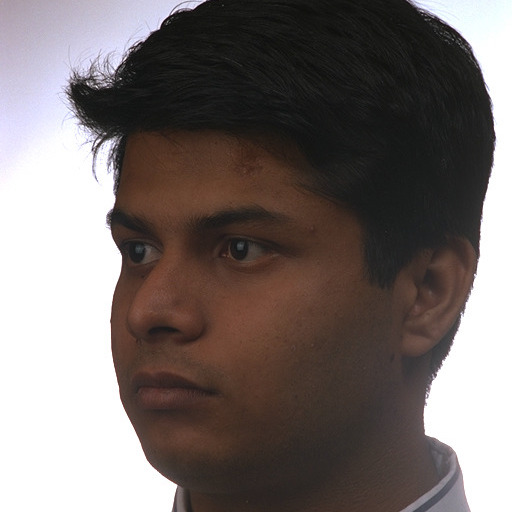} &
\includegraphics[width=0.45\columnwidth]{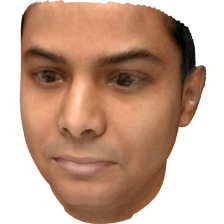}\\

\includegraphics[width=0.45\columnwidth]{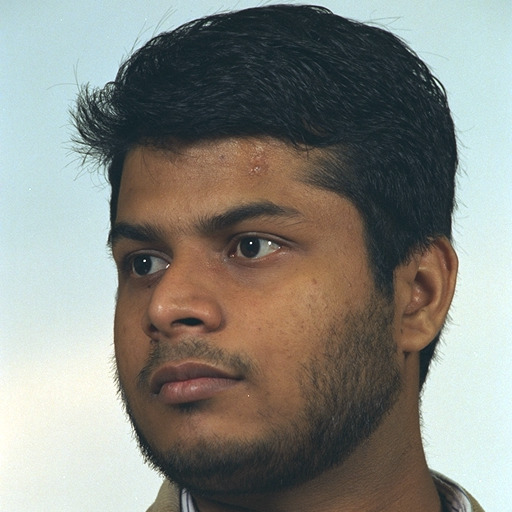} &

\includegraphics[width=0.45\columnwidth]{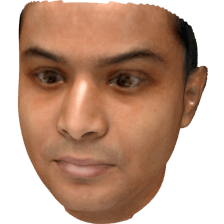}\\

\end{tabular}
\caption{Stability under variable lighting conditions for a subject from the FERET~\cite{feret} dataset.}
\label{fig:lighting}
\end{figure}
% \begin{figure}

% \begin{tabular}{ccc}
% % Our revision: bl10
% %Input & This Paper\\% & Tran et al.\\
% \includegraphics[width=0.3\columnwidth]{pose/input/pose-1.jpg}} &
% \subfloat{\includegraphics[width=0.3\columnwidth]{pose/input/pose-2.jpg}} &
% \subfloat{\includegraphics[width=0.3\columnwidth]{pose/input/pose-3.jpg}}\\

% \subfloat{\includegraphics[width=0.3\columnwidth]{pose/us/pose-1.png}} &
% \subfloat{\includegraphics[width=0.3\columnwidth]{pose/us/pose-2.png}} &
% \subfloat{\includegraphics[width=0.3\columnwidth]{pose/us/pose-3.png}}\\

% \end{tabular}
% \caption{Pose Stress Test on a subject from the FERET~\cite{feret} dataset. Our algorithm is consistent under a $45^\circ$ rotation. Under a $90^\circ$ rotation, global shape changes.}
% \label{fig:pose}
% \end{figure}

\begin{figure}[p]
\vspace{-7.6cm}
\begin{tabular}{cc}

% Our revision: bl10
%Input & This Paper\\% & Tran et al.\\
\includegraphics[width=0.45\columnwidth]{pose/input/pose-1.jpg} &
\includegraphics[width=0.45\columnwidth]{pose/us/pose-1.png}\\

\includegraphics[width=0.45\columnwidth]{pose/input/pose-2.jpg} &
\includegraphics[width=0.45\columnwidth]{pose/us/pose-2.png}\\

\includegraphics[width=0.45\columnwidth]{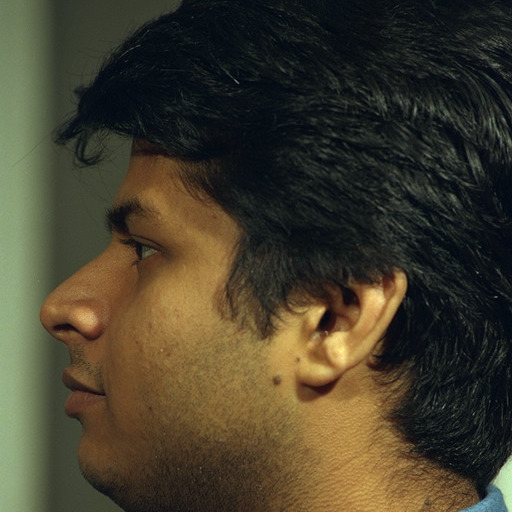} &
\includegraphics[width=0.45\columnwidth]{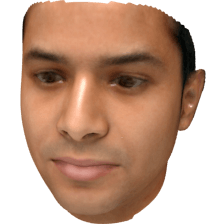}\\

\end{tabular}
\caption{Pose Stress Test on a subject from the FERET~\cite{feret} dataset. Our algorithm is consistent under a $45^\circ$ rotation. Under a $90^\circ$ rotation, global shape changes.}
\label{fig:pose}
\end{figure}
\begin{figure*}
\begin{tabular}{cccc}
% Inputs
% Revision: bl10
\includegraphics[width = 0.23\textwidth]{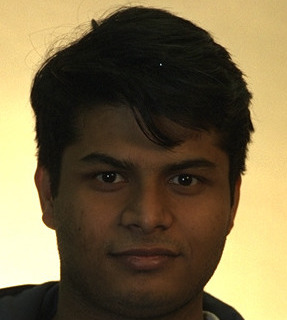} &
\includegraphics[width = 0.23\textwidth]{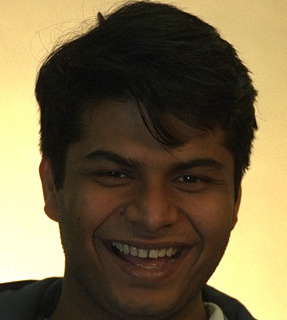} &
\includegraphics[width = 0.23\textwidth]{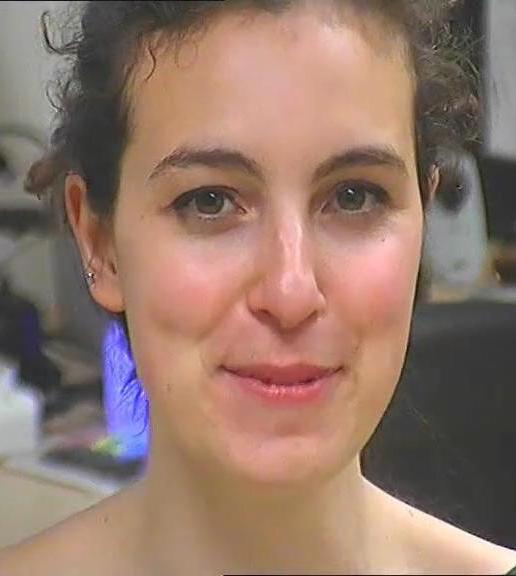} &
\includegraphics[width = 0.23\textwidth]{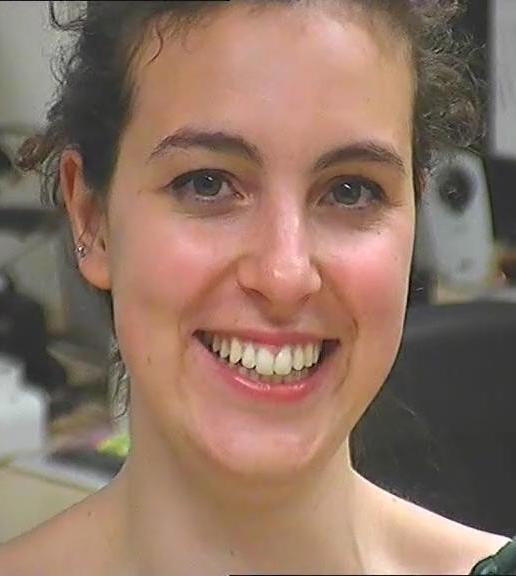}\\
% Us, revision bl10
\includegraphics[width = 0.23\textwidth]{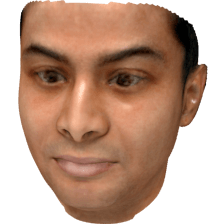} &
\includegraphics[width = 0.23\textwidth]{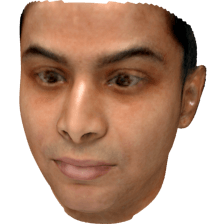} &
\includegraphics[width = 0.23\textwidth]{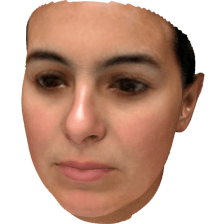} &
\includegraphics[width = 0.23\textwidth]{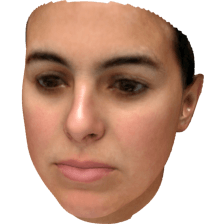}\\
\end{tabular}
\caption{Expression stability test on subjects from the FERET~\cite{feret} (left) and MICC~\cite{Bagdanov:2011:FHF:2072572.2072597} (right) datasets.
For both subjects, our method is invariant to expression, while remaining sensitive to identity.}
\label{fig:expression}
\end{figure*}
\begin{figure*}
\begin{tabular}{ccccc}
% Inputs
% Revision: bl10
\includegraphics[width=0.177\textwidth]{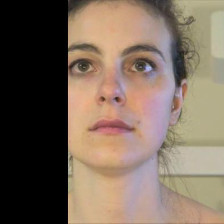} &
\includegraphics[width=0.177\textwidth]{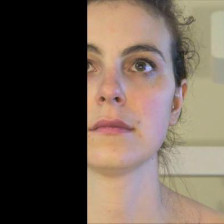} &
\includegraphics[width=0.177\textwidth]{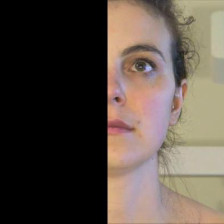} &
\includegraphics[width=0.177\textwidth]{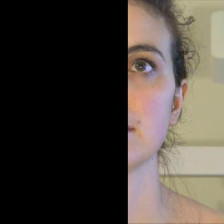} &
\includegraphics[width=0.177\textwidth]{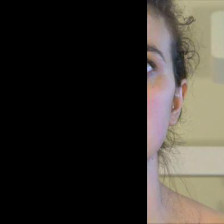}\\
% Us
\includegraphics[width=0.177\textwidth]{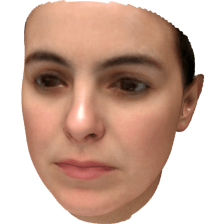} &
\includegraphics[width=0.177\textwidth]{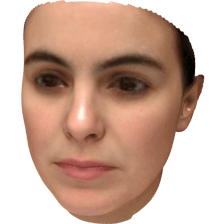} &
\includegraphics[width=0.177\textwidth]{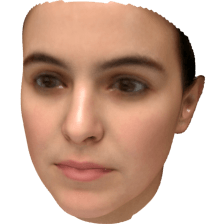} &
\includegraphics[width=0.177\textwidth]{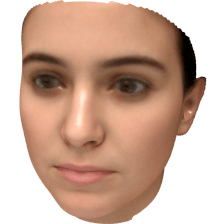} &
\includegraphics[width=0.177\textwidth]{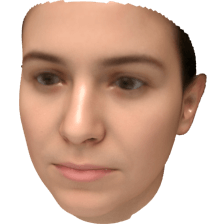}\\

\includegraphics[width=0.177\textwidth]{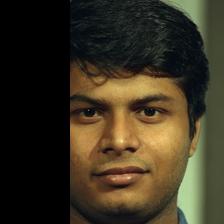} &
\includegraphics[width=0.177\textwidth]{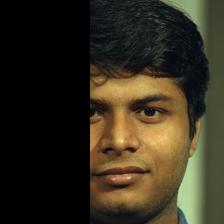} &
\includegraphics[width=0.177\textwidth]{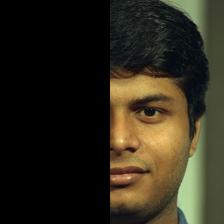} &
\includegraphics[width=0.177\textwidth]{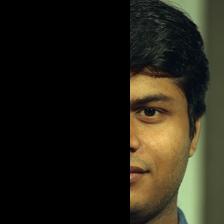} &
\includegraphics[width=0.177\textwidth]{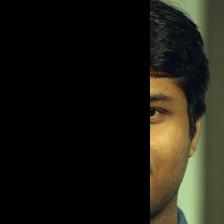}\\

\includegraphics[width=0.177\textwidth]{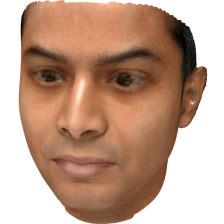} &
\includegraphics[width=0.177\textwidth]{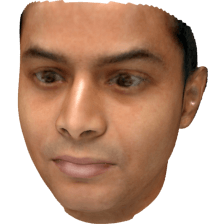} &
\includegraphics[width=0.177\textwidth]{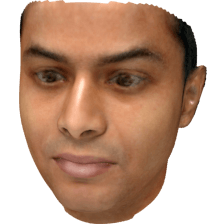} &
\includegraphics[width=0.177\textwidth]{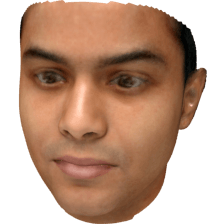} &
\includegraphics[width=0.177\textwidth]{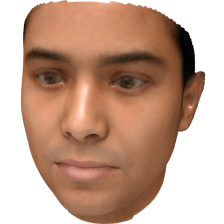}\\
\end{tabular}
\caption{Occlusion Stress Test on subjects from the MICC~\cite{Bagdanov:2011:FHF:2072572.2072597} and FERET~\cite{feret} dataset. We increase occlusion in the input image until our algorithm no longer predicts accurate features. Facial features smoothly degrade as the necessary information is no longer present in the input image.}
\end{figure*}

\begin{figure*}
\begin{tabular}{cccc}
% Inputs
% Revision: bl10
\includegraphics[width = 0.23\textwidth]{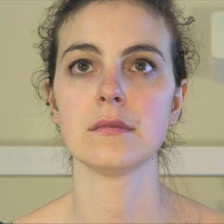} &
\includegraphics[width = 0.23\textwidth]{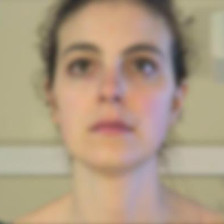} &
\includegraphics[width = 0.23\textwidth]{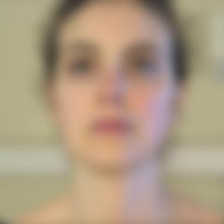} &
\includegraphics[width = 0.23\textwidth]{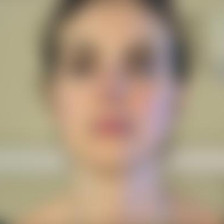}\\

% Us
\includegraphics[width = 0.23\textwidth]{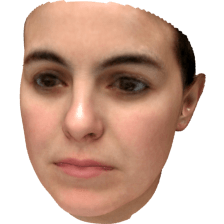} &
\includegraphics[width = 0.23\textwidth]{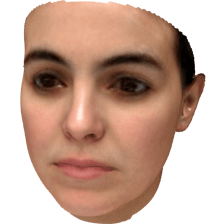} &
\includegraphics[width = 0.23\textwidth]{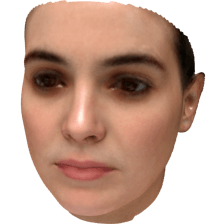} &
\includegraphics[width = 0.23\textwidth]{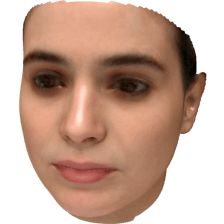}\\

\includegraphics[width = 0.23\textwidth]{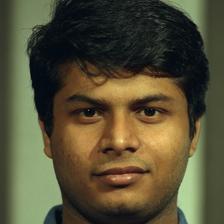} &
\includegraphics[width = 0.23\textwidth]{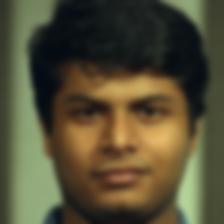} &
\includegraphics[width = 0.23\textwidth]{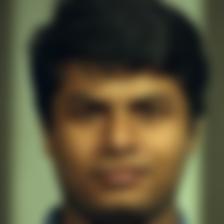} &
\includegraphics[width = 0.23\textwidth]{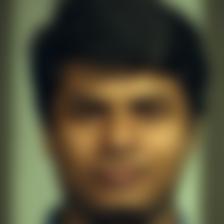}\\

\includegraphics[width = 0.23\textwidth]{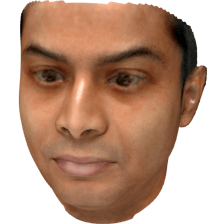} &
\includegraphics[width = 0.23\textwidth]{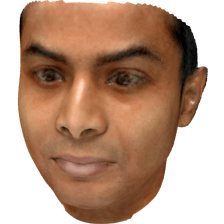} &
\includegraphics[width = 0.23\textwidth]{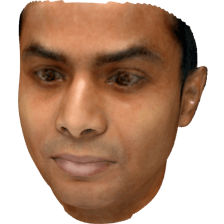} &
\includegraphics[width = 0.23\textwidth]{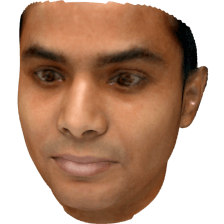}\\
\end{tabular}
\caption{Resolution Stress Test on subjects from the MICC~\cite{Bagdanov:2011:FHF:2072572.2072597} (top) and FERET~\cite{feret} (bottom) datasets. Beginning with a frontal image of the subject, we apply a progressively larger gaussian blur kernel to examine the effect of lost detail in the input. For the female subject, global shape begins to change subtly as the blur becomes extreme. For both subjects, fine detail in the eyebrow shape and thickness is lost as the input is increasingly blurred.}
\end{figure*}
\begin{figure*}
\begin{tabular}{cccccc}
% Inputs
% Revision: bl10
\includegraphics[width=0.14375\textwidth]{banner/Al_Gore_0001.jpg} &
\includegraphics[width=0.14375\textwidth]{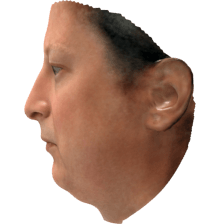} &
\includegraphics[width=0.14375\textwidth]{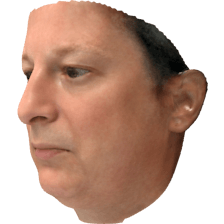} &
\includegraphics[width=0.14375\textwidth]{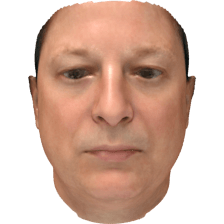} &
\includegraphics[width=0.14375\textwidth]{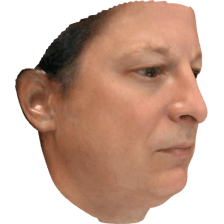} &
\includegraphics[width=0.14375\textwidth]{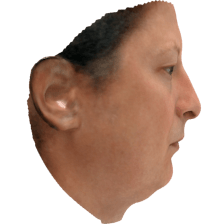}\\

\includegraphics[width=0.14375\textwidth]{banner/Cecilia_Cheung_0001.jpg} &
\includegraphics[width=0.14375\textwidth]{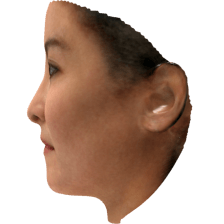} &
\includegraphics[width=0.14375\textwidth]{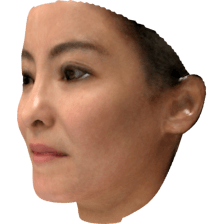} &
\includegraphics[width=0.14375\textwidth]{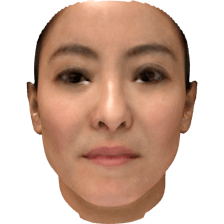} &
\includegraphics[width=0.14375\textwidth]{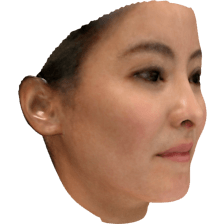} &
\includegraphics[width=0.14375\textwidth]{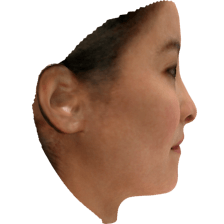}\\

\includegraphics[width=0.14375\textwidth]{banner/Clint_Howard_0001.jpg} &
\includegraphics[width=0.14375\textwidth]{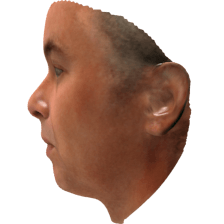} &
\includegraphics[width=0.14375\textwidth]{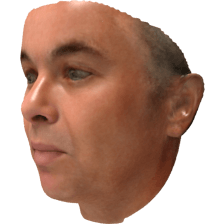} &
\includegraphics[width=0.14375\textwidth]{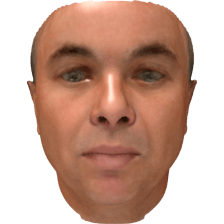} &
\includegraphics[width=0.14375\textwidth]{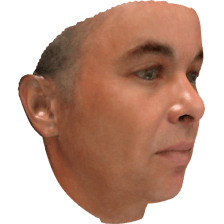} &
\includegraphics[width=0.14375\textwidth]{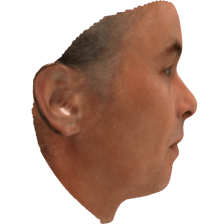}\\

\includegraphics[width=0.14375\textwidth]{banner/Jennifer_Lopez_0003.jpg} &
\includegraphics[width=0.14375\textwidth]{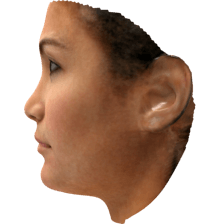} &
\includegraphics[width=0.14375\textwidth]{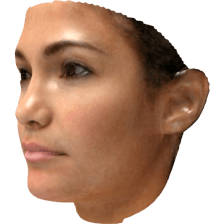} &
\includegraphics[width=0.14375\textwidth]{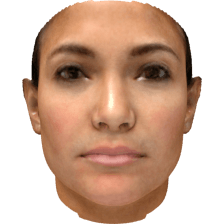} &
\includegraphics[width=0.14375\textwidth]{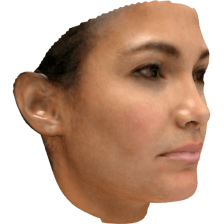} &
\includegraphics[width=0.14375\textwidth]{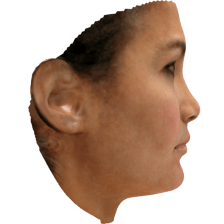}\\

\includegraphics[width=0.14375\textwidth]{banner/Peter_OToole_0001.jpg} &
\includegraphics[width=0.14375\textwidth]{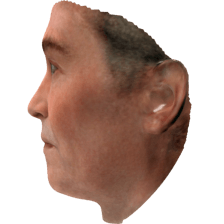} &
\includegraphics[width=0.14375\textwidth]{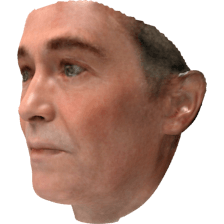} &
\includegraphics[width=0.14375\textwidth]{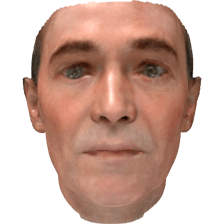} &
\includegraphics[width=0.14375\textwidth]{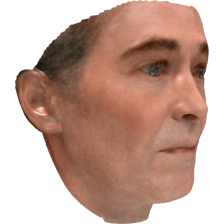} &
\includegraphics[width=0.14375\textwidth]{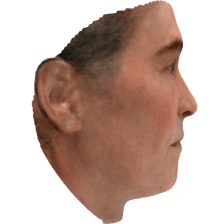}\\

\includegraphics[width=0.14375\textwidth]{banner/Queen_Latifah_0003.jpg} &
\includegraphics[width=0.14375\textwidth]{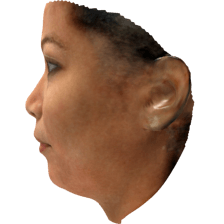} &
\includegraphics[width=0.14375\textwidth]{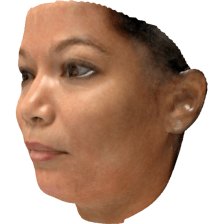} &
\includegraphics[width=0.14375\textwidth]{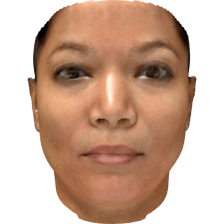} &
\includegraphics[width=0.14375\textwidth]{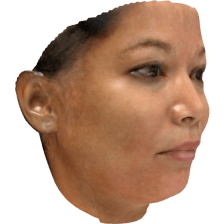} &
\includegraphics[width=0.14375\textwidth]{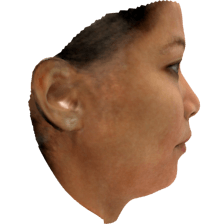}\\
% Us
% \includegraphics[width=0.14375\textwidth]{occlusion/us/2633_11.png}} &
% \includegraphics[width=0.14375\textwidth]{occlusion/us/2633_13.png}} &
% \includegraphics[width=0.14375\textwidth]{occlusion/us/2633_15.png}} &
% \includegraphics[width=0.14375\textwidth]{occlusion/us/2633_17.png}} &
% \includegraphics[width=0.14375\textwidth]{occlusion/us/2633_19.png}}\\

% \includegraphics[width=0.14375\textwidth]{occlusion/feret/in/occl-70.jpg}} &
% \includegraphics[width=0.14375\textwidth]{occlusion/feret/in/occl-90.jpg}} &
% \includegraphics[width=0.14375\textwidth]{occlusion/feret/in/occl-110.jpg}} &
% \includegraphics[width=0.14375\textwidth]{occlusion/feret/in/occl-130.jpg}} &
% \includegraphics[width=0.14375\textwidth]{occlusion/feret/in/occl-150.jpg}}\\

% \includegraphics[width=0.14375\textwidth]{occlusion/feret/us/occl-70.png}} &
% \includegraphics[width=0.14375\textwidth]{occlusion/feret/us/occl-90.png}} &
% \includegraphics[width=0.14375\textwidth]{occlusion/feret/us/occl-110.png}} &
% \includegraphics[width=0.14375\textwidth]{occlusion/feret/us/occl-130.png}} &
% \includegraphics[width=0.14375\textwidth]{occlusion/feret/us/occl-150.png}}\\
\end{tabular}
\caption{Views of the six teaser LFW~\cite{LFWTech} subjects at $-90^{\circ}$, $-45^\circ$, $0^\circ$, $45^\circ$, and $90^\circ$ rotations.}
\end{figure*}

\begin{figure*}
\vspace*{-2cm}
\begin{tabular}{cccccccc}
Input & Ours & Tran\cite{tran2017regressing} & MoFA\cite{tewari17MoFA} & Input & Ours & Tran\cite{tran2017regressing} & MoFA\cite{tewari17MoFA}\\
\includegraphics[width = 0.715in]{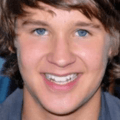} &
\includegraphics[width = 0.715in]{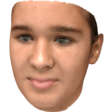} &
\includegraphics[width = 0.715in]{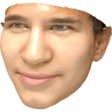} &
\includegraphics[width = 0.715in]{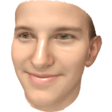} &
\includegraphics[width = 0.715in]{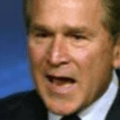} &
\includegraphics[width = 0.715in]{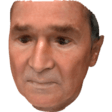} &
\includegraphics[width = 0.715in]{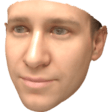} &
\includegraphics[width = 0.715in]{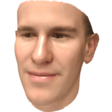}\\

\includegraphics[width = 0.715in]{full-qual/in/02.png} &
\includegraphics[width = 0.715in]{full-qual/us/02.png} &
\includegraphics[width = 0.715in]{full-qual/tran/02.png} &
\includegraphics[width = 0.715in]{full-qual/mofa/02.png} &
\includegraphics[width = 0.715in]{full-qual/in/03.png} &
\includegraphics[width = 0.715in]{full-qual/us/03.png} &
\includegraphics[width = 0.715in]{full-qual/tran/03.png} &
\includegraphics[width = 0.715in]{full-qual/mofa/03.png}\\

\includegraphics[width = 0.715in]{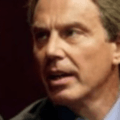} &
\includegraphics[width = 0.715in]{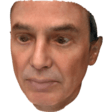} &
\includegraphics[width = 0.715in]{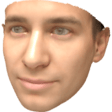} &
\includegraphics[width = 0.715in]{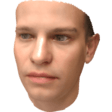} &
\includegraphics[width = 0.715in]{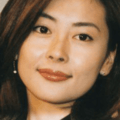} &
\includegraphics[width = 0.715in]{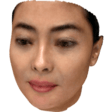} &
\includegraphics[width = 0.715in]{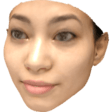} &
\includegraphics[width = 0.715in]{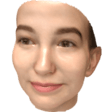}\\

\includegraphics[width = 0.715in]{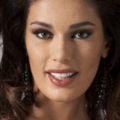} &
\includegraphics[width = 0.715in]{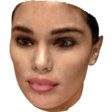} &
\includegraphics[width = 0.715in]{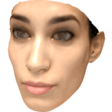} &
\includegraphics[width = 0.715in]{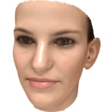} &
\includegraphics[width = 0.715in]{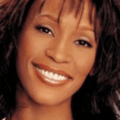} &
\includegraphics[width = 0.715in]{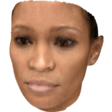} &
\includegraphics[width = 0.715in]{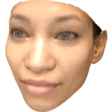} &
\includegraphics[width = 0.715in]{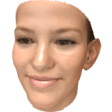}\\

\includegraphics[width = 0.715in]{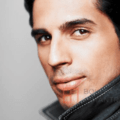} &
\includegraphics[width = 0.715in]{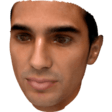} &
\includegraphics[width = 0.715in]{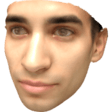} &
\includegraphics[width = 0.715in]{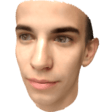} &
\includegraphics[width = 0.715in]{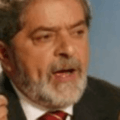} &
\includegraphics[width = 0.715in]{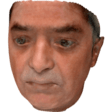} &
\includegraphics[width = 0.715in]{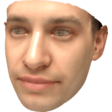} &
\includegraphics[width = 0.715in]{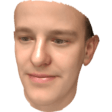}\\

\includegraphics[width = 0.715in]{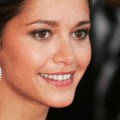} &
\includegraphics[width = 0.715in]{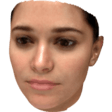} &
\includegraphics[width = 0.715in]{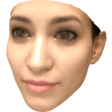} &
\includegraphics[width = 0.715in]{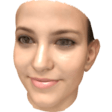} &
\includegraphics[width = 0.715in]{full-qual/in/11.png} &
\includegraphics[width = 0.715in]{full-qual/us/11.png} &
\includegraphics[width = 0.715in]{full-qual/tran/11.png} &
\includegraphics[width = 0.715in]{full-qual/mofa/11.png}\\

\includegraphics[width = 0.715in]{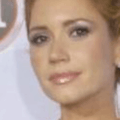} &
\includegraphics[width = 0.715in]{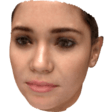} &
\includegraphics[width = 0.715in]{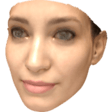} &
\includegraphics[width = 0.715in]{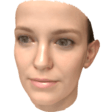} &
\includegraphics[width = 0.715in]{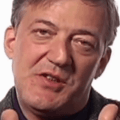} &
\includegraphics[width = 0.715in]{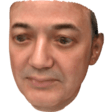} &
\includegraphics[width = 0.715in]{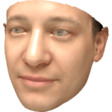} &
\includegraphics[width = 0.715in]{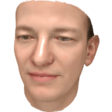}\\

\includegraphics[width = 0.715in]{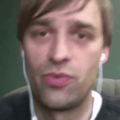} &
\includegraphics[width = 0.715in]{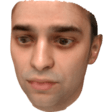} &
\includegraphics[width = 0.715in]{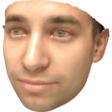} &
\includegraphics[width = 0.715in]{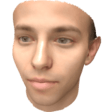} &
\includegraphics[width = 0.715in]{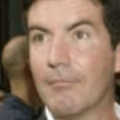} &
\includegraphics[width = 0.715in]{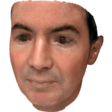} &
\includegraphics[width = 0.715in]{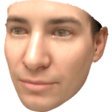} &
\includegraphics[width = 0.715in]{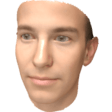}\\

\includegraphics[width = 0.715in]{full-qual/in/16.png} &
\includegraphics[width = 0.715in]{full-qual/us/16.png} &
\includegraphics[width = 0.715in]{full-qual/tran/16.png} &
\includegraphics[width = 0.715in]{full-qual/mofa/16.png} &
\includegraphics[width = 0.715in]{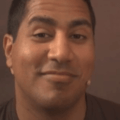} &
\includegraphics[width = 0.715in]{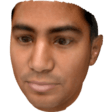} &
\includegraphics[width = 0.715in]{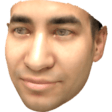} &
\includegraphics[width = 0.715in]{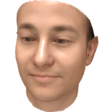}\\

\includegraphics[width = 0.715in]{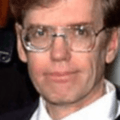} &
\includegraphics[width = 0.715in]{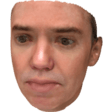} &
\includegraphics[width = 0.715in]{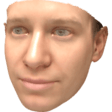} &
\includegraphics[width = 0.715in]{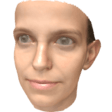} &
\includegraphics[width = 0.715in]{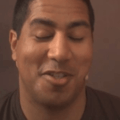} &
\includegraphics[width = 0.715in]{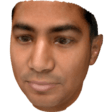} &
\includegraphics[width = 0.715in]{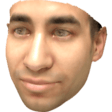} &
\includegraphics[width = 0.715in]{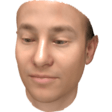}\\

\includegraphics[width = 0.715in]{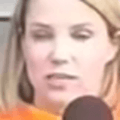} &
\includegraphics[width = 0.715in]{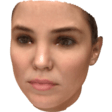} &
\includegraphics[width = 0.715in]{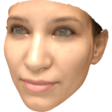} &
\includegraphics[width = 0.715in]{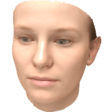} &
\includegraphics[width = 0.715in]{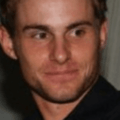} &
\includegraphics[width = 0.715in]{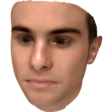} &
\includegraphics[width = 0.715in]{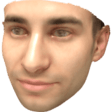} &
\includegraphics[width = 0.715in]{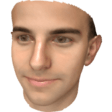}\\

\end{tabular}
\caption{Full qualitative comparison on the MoFA-Test dataset. Results 1-22.}
\label{fig:full-qual-1}
\end{figure*}

\begin{figure*}
\vspace*{-2cm}
\begin{tabular}{cccccccc}
Input & Ours & Tran\cite{tran2017regressing} & MoFA\cite{tewari17MoFA} & Input & Ours & Tran\cite{tran2017regressing} & MoFA\cite{tewari17MoFA}\\
\includegraphics[width = 0.715in]{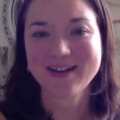} &
\includegraphics[width = 0.715in]{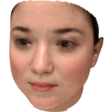} &
\includegraphics[width = 0.715in]{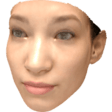} &
\includegraphics[width = 0.715in]{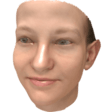} &
\includegraphics[width = 0.715in]{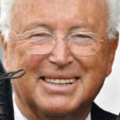} &
\includegraphics[width = 0.715in]{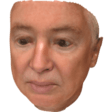} &
\includegraphics[width = 0.715in]{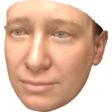} &
\includegraphics[width = 0.715in]{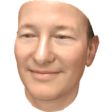}\\

\includegraphics[width = 0.715in]{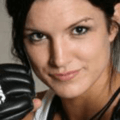} &
\includegraphics[width = 0.715in]{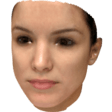} &
\includegraphics[width = 0.715in]{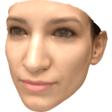} &
\includegraphics[width = 0.715in]{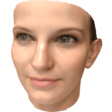} &
\includegraphics[width = 0.715in]{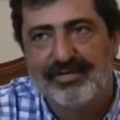} &
\includegraphics[width = 0.715in]{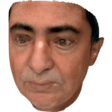} &
\includegraphics[width = 0.715in]{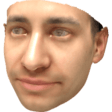} &
\includegraphics[width = 0.715in]{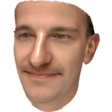}\\

\includegraphics[width = 0.715in]{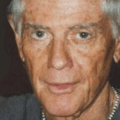} &
\includegraphics[width = 0.715in]{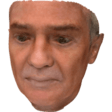} &
\includegraphics[width = 0.715in]{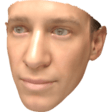} &
\includegraphics[width = 0.715in]{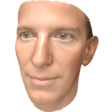} &
\includegraphics[width = 0.715in]{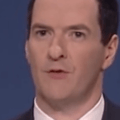} &
\includegraphics[width = 0.715in]{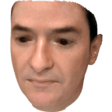} &
\includegraphics[width = 0.715in]{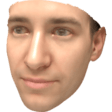} &
\includegraphics[width = 0.715in]{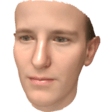}\\

\includegraphics[width = 0.715in]{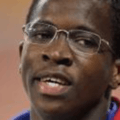} &
\includegraphics[width = 0.715in]{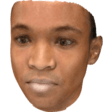} &
\includegraphics[width = 0.715in]{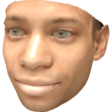} &
\includegraphics[width = 0.715in]{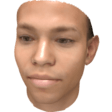} &
\includegraphics[width = 0.715in]{full-qual/in/29.png} &
\includegraphics[width = 0.715in]{full-qual/us/29.png} &
\includegraphics[width = 0.715in]{full-qual/tran/29.png} &
\includegraphics[width = 0.715in]{full-qual/mofa/29.png}\\

\includegraphics[width = 0.715in]{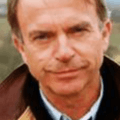} &
\includegraphics[width = 0.715in]{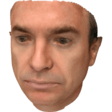} &
\includegraphics[width = 0.715in]{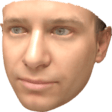} &
\includegraphics[width = 0.715in]{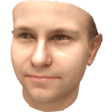} &
\includegraphics[width = 0.715in]{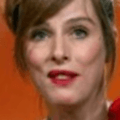} &
\includegraphics[width = 0.715in]{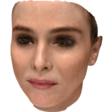} &
\includegraphics[width = 0.715in]{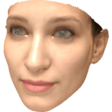} &
\includegraphics[width = 0.715in]{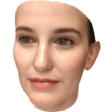}\\

\includegraphics[width = 0.715in]{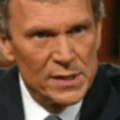} &
\includegraphics[width = 0.715in]{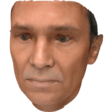} &
\includegraphics[width = 0.715in]{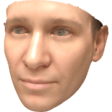} &
\includegraphics[width = 0.715in]{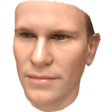} &
\includegraphics[width = 0.715in]{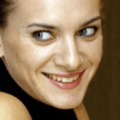} &
\includegraphics[width = 0.715in]{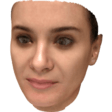} &
\includegraphics[width = 0.715in]{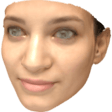} &
\includegraphics[width = 0.715in]{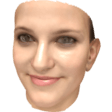}\\

\includegraphics[width = 0.715in]{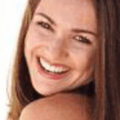} &
\includegraphics[width = 0.715in]{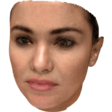} &
\includegraphics[width = 0.715in]{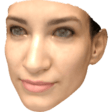} &
\includegraphics[width = 0.715in]{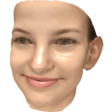} &
\includegraphics[width = 0.715in]{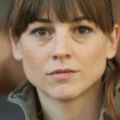} &
\includegraphics[width = 0.715in]{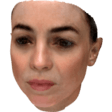} &
\includegraphics[width = 0.715in]{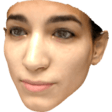} &
\includegraphics[width = 0.715in]{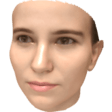}\\

\includegraphics[width = 0.715in]{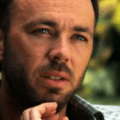} &
\includegraphics[width = 0.715in]{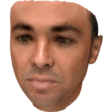} &
\includegraphics[width = 0.715in]{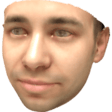} &
\includegraphics[width = 0.715in]{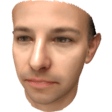} &
\includegraphics[width = 0.715in]{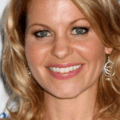} &
\includegraphics[width = 0.715in]{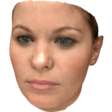} &
\includegraphics[width = 0.715in]{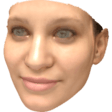} &
\includegraphics[width = 0.715in]{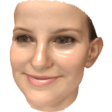}\\

\includegraphics[width = 0.715in]{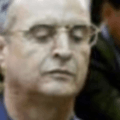} &
\includegraphics[width = 0.715in]{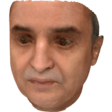} &
\includegraphics[width = 0.715in]{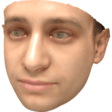} &
\includegraphics[width = 0.715in]{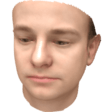} &
\includegraphics[width = 0.715in]{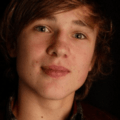} &
\includegraphics[width = 0.715in]{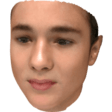} &
\includegraphics[width = 0.715in]{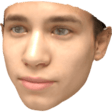} &
\includegraphics[width = 0.715in]{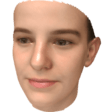}\\

\includegraphics[width = 0.715in]{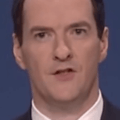} &
\includegraphics[width = 0.715in]{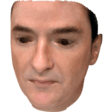} &
\includegraphics[width = 0.715in]{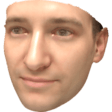} &
\includegraphics[width = 0.715in]{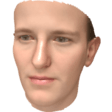} &
\includegraphics[width = 0.715in]{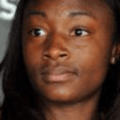} &
\includegraphics[width = 0.715in]{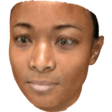} &
\includegraphics[width = 0.715in]{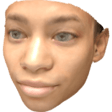} &
\includegraphics[width = 0.715in]{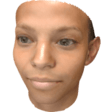}\\

\includegraphics[width = 0.715in]{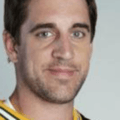} &
\includegraphics[width = 0.715in]{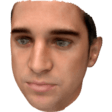} &
\includegraphics[width = 0.715in]{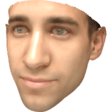} &
\includegraphics[width = 0.715in]{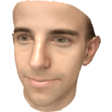} &
\includegraphics[width = 0.715in]{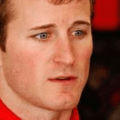} &
\includegraphics[width = 0.715in]{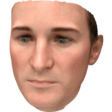} &
\includegraphics[width = 0.715in]{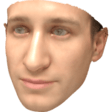} &
\includegraphics[width = 0.715in]{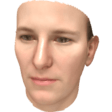}\\

\end{tabular}
\caption{Full qualitative comparison on the MoFA-Test dataset. Results 23-44.}
\label{fig:full-qual-2}
\end{figure*}

\begin{figure*}
\vspace*{-2cm}
\begin{tabular}{cccccccc}
Input & Ours & Tran\cite{tran2017regressing} & MoFA\cite{tewari17MoFA} & Input & Ours & Tran\cite{tran2017regressing} & MoFA\cite{tewari17MoFA}\\
\includegraphics[width = 0.715in]{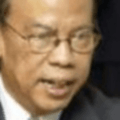} &
\includegraphics[width = 0.715in]{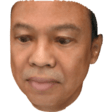} &
\includegraphics[width = 0.715in]{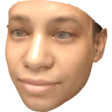} &
\includegraphics[width = 0.715in]{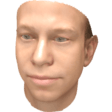} &
\includegraphics[width = 0.715in]{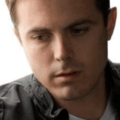} &
\includegraphics[width = 0.715in]{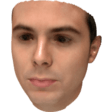} &
\includegraphics[width = 0.715in]{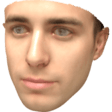} &
\includegraphics[width = 0.715in]{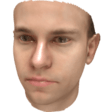}\\

\includegraphics[width = 0.715in]{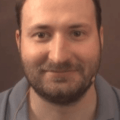} &
\includegraphics[width = 0.715in]{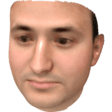} &
\includegraphics[width = 0.715in]{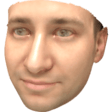} &
\includegraphics[width = 0.715in]{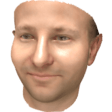} &
\includegraphics[width = 0.715in]{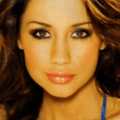} &
\includegraphics[width = 0.715in]{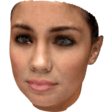} &
\includegraphics[width = 0.715in]{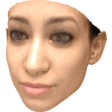} &
\includegraphics[width = 0.715in]{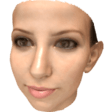}\\

\includegraphics[width = 0.715in]{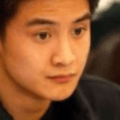} &
\includegraphics[width = 0.715in]{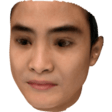} &
\includegraphics[width = 0.715in]{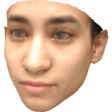} &
\includegraphics[width = 0.715in]{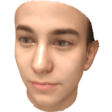} &
\includegraphics[width = 0.715in]{full-qual/in/49.png} &
\includegraphics[width = 0.715in]{full-qual/us/49.png} &
\includegraphics[width = 0.715in]{full-qual/tran/49.png} &
\includegraphics[width = 0.715in]{full-qual/mofa/49.png}\\

\includegraphics[width = 0.715in]{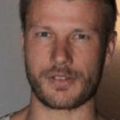} &
\includegraphics[width = 0.715in]{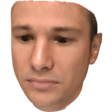} &
\includegraphics[width = 0.715in]{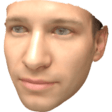} &
\includegraphics[width = 0.715in]{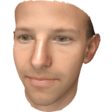} &
\includegraphics[width = 0.715in]{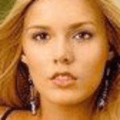} &
\includegraphics[width = 0.715in]{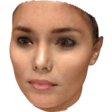} &
\includegraphics[width = 0.715in]{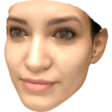} &
\includegraphics[width = 0.715in]{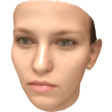}\\

\includegraphics[width = 0.715in]{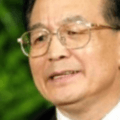} &
\includegraphics[width = 0.715in]{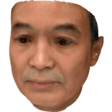} &
\includegraphics[width = 0.715in]{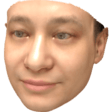} &
\includegraphics[width = 0.715in]{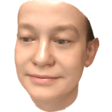} &
\includegraphics[width = 0.715in]{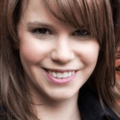} &
\includegraphics[width = 0.715in]{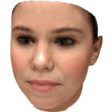} &
\includegraphics[width = 0.715in]{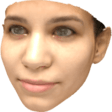} &
\includegraphics[width = 0.715in]{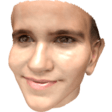}\\

\includegraphics[width = 0.715in]{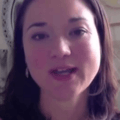} &
\includegraphics[width = 0.715in]{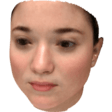} &
\includegraphics[width = 0.715in]{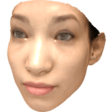} &
\includegraphics[width = 0.715in]{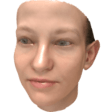} &
\includegraphics[width = 0.715in]{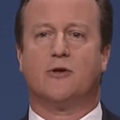} &
\includegraphics[width = 0.715in]{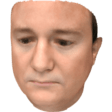} &
\includegraphics[width = 0.715in]{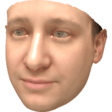} &
\includegraphics[width = 0.715in]{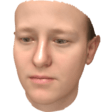}\\

\includegraphics[width = 0.715in]{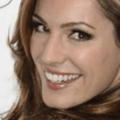} &
\includegraphics[width = 0.715in]{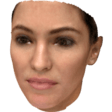} &
\includegraphics[width = 0.715in]{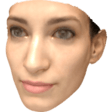} &
\includegraphics[width = 0.715in]{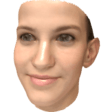} &
\includegraphics[width = 0.715in]{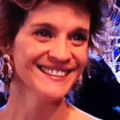} &
\includegraphics[width = 0.715in]{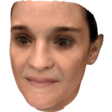} &
\includegraphics[width = 0.715in]{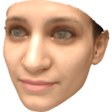} &
\includegraphics[width = 0.715in]{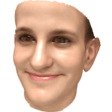}\\

\includegraphics[width = 0.715in]{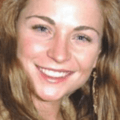} &
\includegraphics[width = 0.715in]{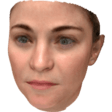} &
\includegraphics[width = 0.715in]{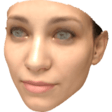} &
\includegraphics[width = 0.715in]{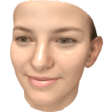} &
\includegraphics[width = 0.715in]{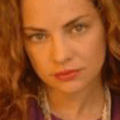} &
\includegraphics[width = 0.715in]{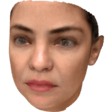} &
\includegraphics[width = 0.715in]{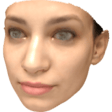} &
\includegraphics[width = 0.715in]{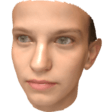}\\

\includegraphics[width = 0.715in]{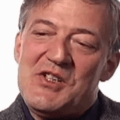} &
\includegraphics[width = 0.715in]{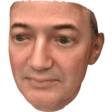} &
\includegraphics[width = 0.715in]{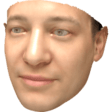} &
\includegraphics[width = 0.715in]{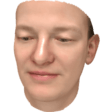} &
\includegraphics[width = 0.715in]{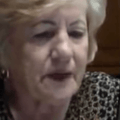} &
\includegraphics[width = 0.715in]{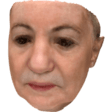} &
\includegraphics[width = 0.715in]{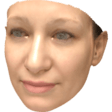} &
\includegraphics[width = 0.715in]{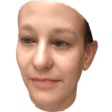}\\

\includegraphics[width = 0.715in]{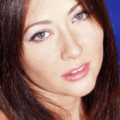} &
\includegraphics[width = 0.715in]{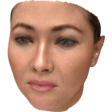} &
\includegraphics[width = 0.715in]{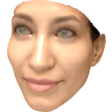} &
\includegraphics[width = 0.715in]{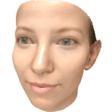} &
\includegraphics[width = 0.715in]{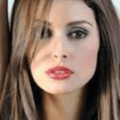} &
\includegraphics[width = 0.715in]{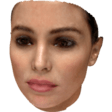} &
\includegraphics[width = 0.715in]{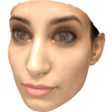} &
\includegraphics[width = 0.715in]{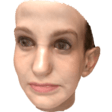}\\

\includegraphics[width = 0.715in]{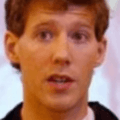} &
\includegraphics[width = 0.715in]{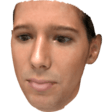} &
\includegraphics[width = 0.715in]{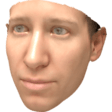} &
\includegraphics[width = 0.715in]{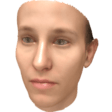} &
\includegraphics[width = 0.715in]{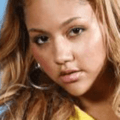} &
\includegraphics[width = 0.715in]{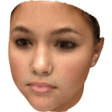} &
\includegraphics[width = 0.715in]{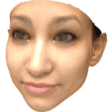} &
\includegraphics[width = 0.715in]{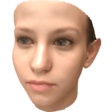}\\

\end{tabular}
\caption{Full qualitative comparison on the MoFA-Test dataset. Results 45-66.}
\label{fig:full-qual-3}
\end{figure*}

\begin{figure*}
\vspace*{-2cm}
\begin{tabular}{cccccccc}
Input & Ours & Tran\cite{tran2017regressing} & MoFA\cite{tewari17MoFA} & Input & Ours & Tran\cite{tran2017regressing} & MoFA\cite{tewari17MoFA}\\
\includegraphics[width = 0.715in]{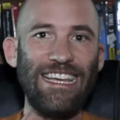} &
\includegraphics[width = 0.715in]{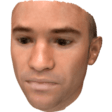} &
\includegraphics[width = 0.715in]{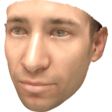} &
\includegraphics[width = 0.715in]{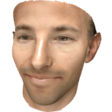} &
\includegraphics[width = 0.715in]{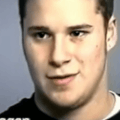} &
\includegraphics[width = 0.715in]{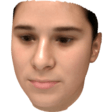} &
\includegraphics[width = 0.715in]{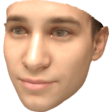} &
\includegraphics[width = 0.715in]{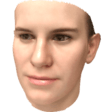}\\

\includegraphics[width = 0.715in]{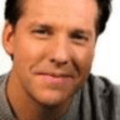} &
\includegraphics[width = 0.715in]{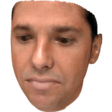} &
\includegraphics[width = 0.715in]{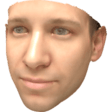} &
\includegraphics[width = 0.715in]{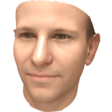} &
\includegraphics[width = 0.715in]{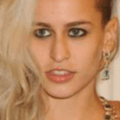} &
\includegraphics[width = 0.715in]{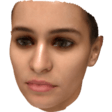} &
\includegraphics[width = 0.715in]{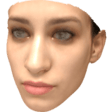} &
\includegraphics[width = 0.715in]{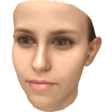}\\

\includegraphics[width = 0.715in]{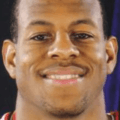} &
\includegraphics[width = 0.715in]{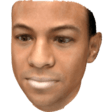} &
\includegraphics[width = 0.715in]{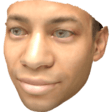} &
\includegraphics[width = 0.715in]{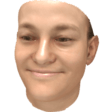} &
\includegraphics[width = 0.715in]{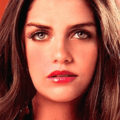} &
\includegraphics[width = 0.715in]{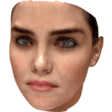} &
\includegraphics[width = 0.715in]{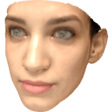} &
\includegraphics[width = 0.715in]{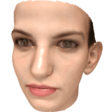}\\

\includegraphics[width = 0.715in]{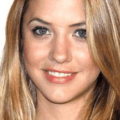} &
\includegraphics[width = 0.715in]{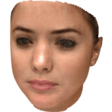} &
\includegraphics[width = 0.715in]{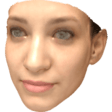} &
\includegraphics[width = 0.715in]{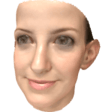} &
\includegraphics[width = 0.715in]{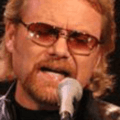} &
\includegraphics[width = 0.715in]{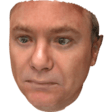} &
\includegraphics[width = 0.715in]{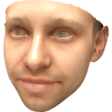} &
\includegraphics[width = 0.715in]{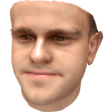}\\

\includegraphics[width = 0.715in]{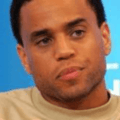} &
\includegraphics[width = 0.715in]{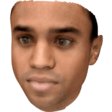} &
\includegraphics[width = 0.715in]{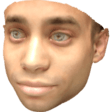} &
\includegraphics[width = 0.715in]{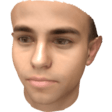} &
\includegraphics[width = 0.715in]{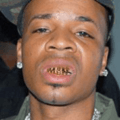} &
\includegraphics[width = 0.715in]{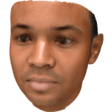} &
\includegraphics[width = 0.715in]{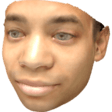} &
\includegraphics[width = 0.715in]{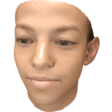}\\

\includegraphics[width = 0.715in]{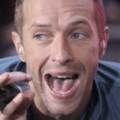} &
\includegraphics[width = 0.715in]{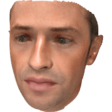} &
\includegraphics[width = 0.715in]{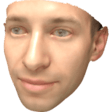} &
\includegraphics[width = 0.715in]{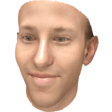} &
\includegraphics[width = 0.715in]{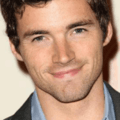} &
\includegraphics[width = 0.715in]{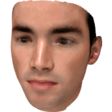} &
\includegraphics[width = 0.715in]{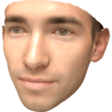} &
\includegraphics[width = 0.715in]{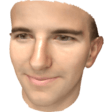}\\

\includegraphics[width = 0.715in]{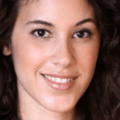} &
\includegraphics[width = 0.715in]{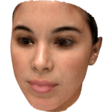} &
\includegraphics[width = 0.715in]{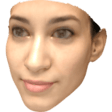} &
\includegraphics[width = 0.715in]{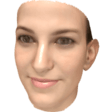} &
\includegraphics[width = 0.715in]{full-qual/in/79.png} &
\includegraphics[width = 0.715in]{full-qual/us/79.png} &
\includegraphics[width = 0.715in]{full-qual/tran/79.png} &
\includegraphics[width = 0.715in]{full-qual/mofa/79.png}\\

\includegraphics[width = 0.715in]{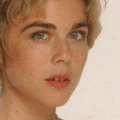} &
\includegraphics[width = 0.715in]{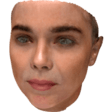} &
\includegraphics[width = 0.715in]{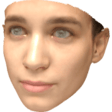} &
\includegraphics[width = 0.715in]{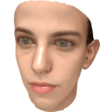} &
\includegraphics[width = 0.715in]{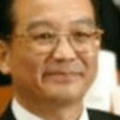} &
\includegraphics[width = 0.715in]{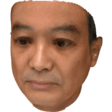} &
\includegraphics[width = 0.715in]{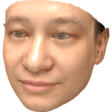} &
\includegraphics[width = 0.715in]{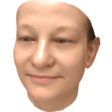}\\

\includegraphics[width = 0.715in]{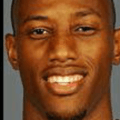} &
\includegraphics[width = 0.715in]{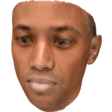} &
\includegraphics[width = 0.715in]{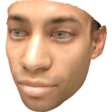} &
\includegraphics[width = 0.715in]{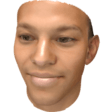} &
\includegraphics[width = 0.715in]{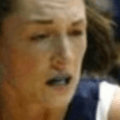} &
\includegraphics[width = 0.715in]{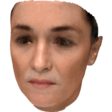} &
\includegraphics[width = 0.715in]{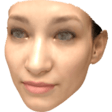} &
\includegraphics[width = 0.715in]{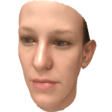}\\

\end{tabular}
\caption{Full qualitative comparison on the MoFA-Test dataset. Results 67-84.}
\label{fig:full-qual-4}
\end{figure*}

\clearpage
\subsection{Fitting Pose and Expression}
\label{sec:landmarks}

\begin{figure}[h]
\begin{tabular}{cc}
\hspace{0.15cm}
\includegraphics[width=0.2\textwidth]{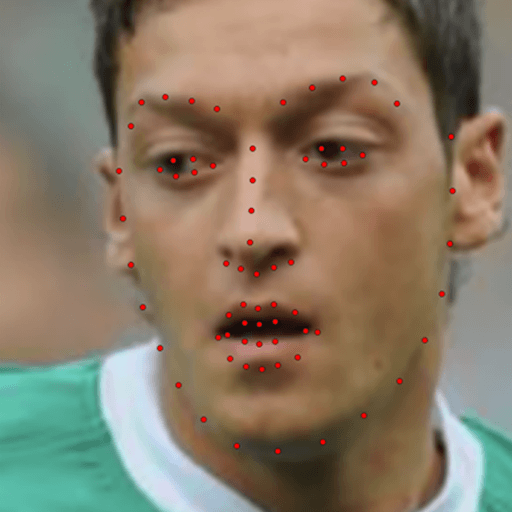} &
\includegraphics[width=0.2\textwidth]{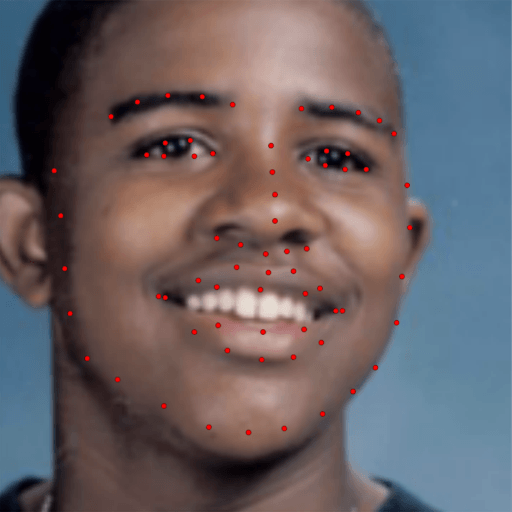}\\

\hspace{0.15cm}
\includegraphics[width=0.2\textwidth]{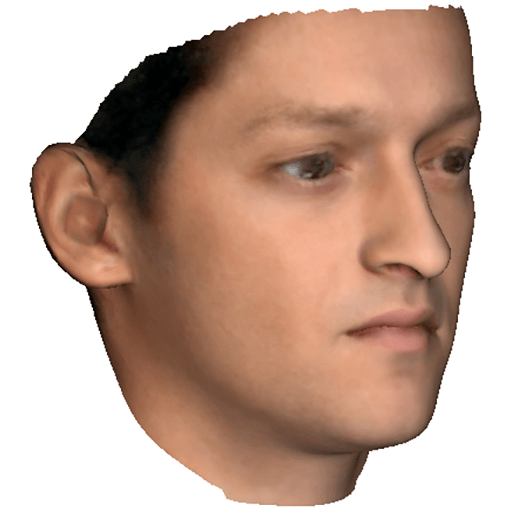} &
\includegraphics[width=0.2\textwidth]{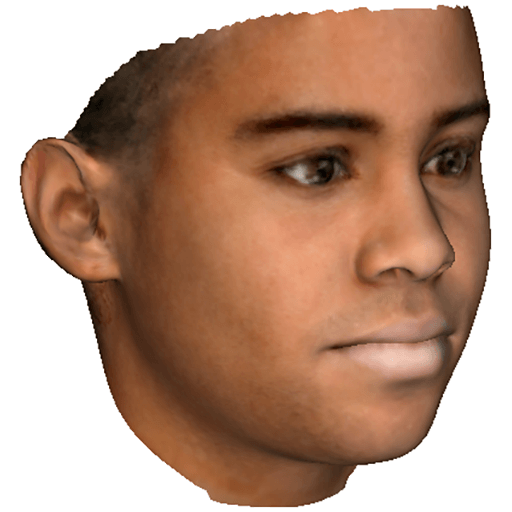}\\

\hspace{0.15cm}
\includegraphics[width=0.2\textwidth]{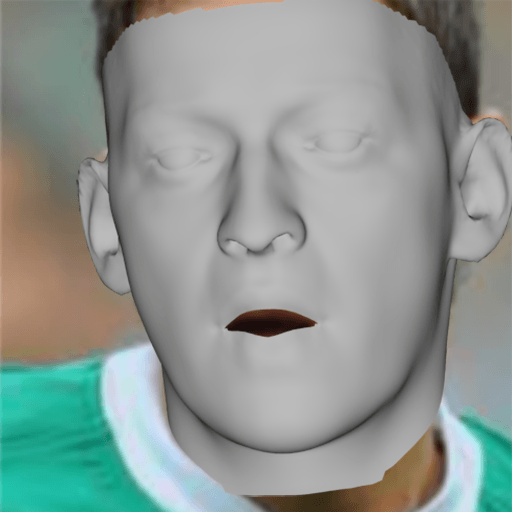} &
\includegraphics[width=0.2\textwidth]{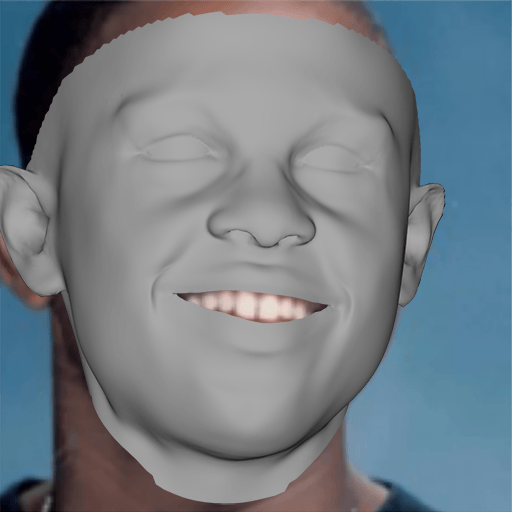}\\

\hspace{0.15cm}
\includegraphics[width=0.2\textwidth]{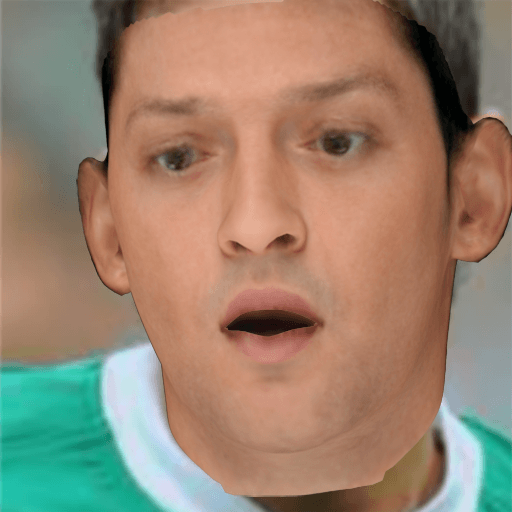} &
\includegraphics[width=0.2\textwidth]{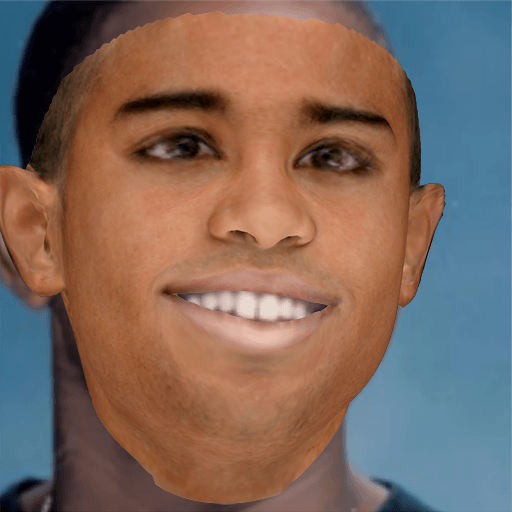} \\

\end{tabular}
\caption{Starting with a neutral face, we used landmarks to fit pose and expression of the morphable model.  Left-to-right: a face image with landmarks, reconstructed neutral face, shaded geometry and albedo overlays with correct pose and expression.}
\label{fig:fitting}
\end{figure}

Our system reconstructs shape and texture of faces, and ignores aspects such as pose, expression, and lighting. Those components are needed to exactly match the reconstruction to the source image, and our neutral face output is an excellent starting point to find them. Figure~\ref{fig:fitting} shows results of gradient descent that starts with our output and fits the pose and expression by minimizing the distances of landmarks on our mesh and the image (we used the 68 landmark configuration from the Multi-PIE database \cite{gross2008mpie}).
% Supplemental Version:

% Paper Version:
% \begin{figure}
% \begin{tabular}{cccc}
% \hspace{-0.3cm}
% \includegraphics[width=0.7in]{fitting/mofa1_markers.png} &
% \includegraphics[width=0.7in]{fitting/mofa1_identity.png} &
% \includegraphics[width=0.7in]{fitting/mofa1_overlay_shaded.png} &
% \includegraphics[width=0.7in]{fitting/mofa1_overlay_colored.png} \\

% \hspace{-0.3cm}
% \includegraphics[width=0.7in]{fitting/mofa3_markers.png} &
% \includegraphics[width=0.7in]{fitting/mofa3_identity.png} &
% \includegraphics[width=0.7in]{fitting/mofa3_overlay_shaded.png} &
% \includegraphics[width=0.7in]{fitting/mofa3_overlay_colored.png} \\

% \end{tabular}
% \caption{Starting with a neutral face, we used landmarks to fit pose and expression of the morphable model.  Left-to-right: a face image with landmarks, reconstructed neutral face, shaded geometry and albedo overlays with correct pose and expression.}
% \label{fig:fitting}
% \end{figure}
\clearpage

\end{document}